\documentclass[sigconf]{acmart}

 \usepackage{amsmath,latexsym,amssymb,amsthm,bbm,array,multirow}

\def\BibTeX{{\rm B\kern-.05em{\sc i\kern-.025em b}\kern-.08emT\kern-.1667em\lower.7ex\hbox{E}\kern-.125emX}}
    
\copyrightyear{2019}
\acmYear{2019}
\setcopyright{acmlicensed}
\acmConference[Woodstock '18]{Woodstock '18: ACM Symposium on Neural Gaze Detection}{June 03--05, 2018}{Woodstock, NY}
\acmBooktitle{Woodstock '18: ACM Symposium on Neural Gaze Detection, June 03--05, 2018, Woodstock, NY}
\acmPrice{15.00}
\acmDOI{10.1145/1122445.1122456}
\acmISBN{978-1-4503-9999-9/18/06}

\begin{document}
\title{A Holistic Approach to Interpretability in Financial Lending: Models, Visualizations, and Summary-Explanations}

\author{Chaofan Chen}
  \affiliation{
  Duke University}
  \email{cfchen@cs.duke.edu}
 \author{Kangcheng Lin}
 \affiliation{Duke University}
 \email{kangcheng.lin@duke.edu}
 \author{Cynthia Rudin}
 \affiliation{Duke University}
 \email{cynthia@cs.duke.edu}
 \author{ Yaron Shaposhnik}
 \affiliation{University of Rochester}
 \email{yaron.shaposhnik@gmail.com}
 \author{Sijia Wang}
 \affiliation{Duke University}
 \email{sijia.wang@duke.edu}
 \author{Tong Wang}
 \affiliation{ University of Iowa}
 \email{tong-wang@uiowa.edu}

\renewcommand{\shortauthors}{C. Chen et al.}

\begin{abstract}


 Lending decisions are usually made with proprietary models that provide minimally acceptable explanations to users. In a future world without such secrecy, what decision support tools would one want to use for justified lending decisions? We propose a framework for such decisions, including a globally interpretable machine learning model, an interactive visualization of it, and several types of summaries and explanations for any given decision. The machine learning model is a \textit{two-layer additive risk model}, which is decomposable into subscales, where each node in the second layer represents a meaningful subscale, and all of the nonlinearities are transparent. Our online visualization tool allows exploration of this model, showing precisely how it came to its conclusion. We provide three types of explanations that are simpler than, but consistent with, the global model: case-based reasoning explanations that use neighboring past cases, a set of features that were the most important for the model's prediction, and summary-explanations that provide a customized sparse explanation for any particular lending decision made by the model. Our framework earned the \textit{FICO recognition award} for the Explainable Machine Learning Challenge, which was the first public challenge in the domain of explainable machine learning.\footnote{
 Authors are listed alphabetically.}
\end{abstract}

\maketitle

\begin{frontmatter}

\title{A Holistic Approach to Interpretability in Financial Lending: Models, Visualizations, and Summary-Explanations}
\author[1]{Chaofan Chen}
\author[2]{Kangcheng Lin}
\author[3]{Cynthia Rudin}
\author[4]{Yaron Shaposhnik}
\author[3]{Sijia Wang}
\author[5]{Tong Wang}

\address[1]{University of Maine}
\address[2]{University of Illinois, Urbana-Champaign}
\address[3]{Duke University}
\address[4]{University of Rochester}
\address[5]{University of Iowa}


\begin{abstract}
Lending decisions are usually made with proprietary models that provide minimally acceptable explanations to users. In a future world without such secrecy, what decision support tools would one want to use for justified lending decisions? This question is timely, since the economy has dramatically shifted due to a pandemic, and a massive number of new loans will be necessary in the short term. We propose a framework for such decisions, including a globally interpretable machine learning model, an interactive visualization of it, and several types of summaries and explanations for any given decision. The machine learning model is a \textit{two-layer additive risk model}, which resembles a two-layer neural network, but is decomposable into subscales. In this model, each node in the first (hidden) layer represents a meaningful subscale model, and all of the nonlinearities are transparent. Our online visualization tool allows exploration of this model, showing precisely how it came to its conclusion. We provide three types of explanations that are simpler than, but consistent with, the global model: case-based reasoning explanations that use neighboring past cases, a set of features that were the most important for the model's prediction, and summary-explanations that provide a customized sparse explanation for any particular lending decision made by the model. Our framework earned the \textit{FICO recognition award} for the Explainable Machine Learning Challenge, which was the first public challenge in the domain of explainable machine learning.\footnote{ Authors are listed alphabetically.}
\end{abstract}

\end{frontmatter}

\section{Introduction}
Lending decisions are important high-stakes decisions that can profoundly affect people's lives. 
This is true in ordinary times but even more so these days as the economy rapidly shifts to an era defined by a pandemic, and whole parts of the modern economy need to shift. Effective and transparent decision-making for new loans is paramount.

In accordance with both legal and ethical principles, financial firms are required to provide clear explanations for why a particular loan application was denied \citep{cfpb}. While many financial firms have developed machine learning models for assessing a client's credit risk, these models are often either complex models, or proprietary, and are thus ``black-boxes'' for loan applicants, who do not know how their score is being computed. Even when model explanations are provided, there can be a gap between the explanations for why a loan is denied and the actual computations of the model, because the explanations may be created \textit{posthoc}.

Let us consider, for instance, the posthoc explanations for credit scores that one might receive from one of the major credit rating agencies. The actual calculation of the credit score is proprietary. As an explanation of the score, one might see a notification such as ``Low Credit Usage, Great job keeping usage under 30\%. The lower the better.'' Simultaneously, the same user might receive a warning such as ``Thin File. Having less than 5 open accounts can make you look somewhat riskier. Some lenders would consider you a \textit{thin file}.'' These warnings seem contradictory. (Should the user open more accounts and use more credit? Or should the user decrease credit usage and continue warnings about the thin file?) These are examples of problems that arise with posthoc explanations of a black box model; two posthoc explanations can contradict each other (even when created from the same global model), and the user is left without an understanding of the global model's actual calculations. An understanding of how the variables actually produce the credit score would alleviate this problem.

Laws such as the General Data Protection Regulation and Equal Credit Opportunities Act \citep{regulation2016general,cfpb}, protect the rights of individuals with respect to algorithmic decisions and provide guarantees such as  ``right to explanation.'' While these laws are developing and becoming more expansive, \textit{their impact could be heightened if decision support tools that include risk models were encouraged to be fully transparent}. In order to determine the potential efficacy of transparent risk modeling tools, we would first need to propose ways to build them. \textit{A new framework for transparent decision-making in lending decisions could potentially encourage the veil of secrecy to be lifted from these models}, and provide an immediate alternative approach to current lending approaches once that veil is lifted.

Thus, in this paper, we ask the following question: \textit{if a lending model is not required to be proprietary, ideally what kind of decision-making tools would be useful for important loan decisions?} We aim to answer this question by providing a holistic approach to obtaining fully transparent and interpretable lending decisions. Our approach is a decision support system that includes the following elements: 


\paragraph{1.) The Global Model} We introduce a novel globally interpretable machine learning model, called a \textit{two-layer additive risk model}, to demonstrate what an ideal transparent credit decision model would look like. 
It resembles a traditional credit-scoring model built on subscale scores, where features are partitioned into meaningful subgroups called subscales, and the subscale scores are later combined to form the global model. 
In fact, our new machine learning approach preserves important elements of classical hand-crafted risk scores (decomposable, uses linear modeling, positive coefficients for risk factors), but inserts (interpretable) nonlinearities in several places to make the model more flexible and accurate. 
Our model is constrained to preserve the monotonic relationship between certain features and the predicted risk.
Our 
model is globally interpretable, in the sense that 
its entire reasoning process is reflected in the full computation of the model, and can be easily explained to humans.

\paragraph{2.) Important Risk Factors} Our global model also comes with a natural way to find important decision factors for why a loan is denied. In particular, we can identify the most important subscales and then the most important risk factors within each subscale.

\paragraph{3.) Summary-Explanations} Even though our global model is globally interpretable and thus can be explained on its own, we can also produce optional local \textit{summaries} of general trends in the global model. These are summaries rather than explanations (or \textit{summary-explanations}) in that they do not aim to reproduce the global model, only to show patterns in its predictions. The summary-explanation method we use (called \texttt{OptConsistentRule}) is a model-agnostic explanation algorithm \citep{ShaposhnikRu18}\footnote{The cited paper was written by two of the coauthors of this paper and not yet peer reviewed.}. 
For instance, for a particular applicant's denied loan, a summary-explanation might say that ``all other 594 past applicants with average months in file below 48 were predicted to default on their loan.'' Summary-explanations can be constructed on a per-applicant basis.

\paragraph{4.) Case-based Explanations} Given a (previously unseen) test case, we can find known similar cases that receive the same predictions from the global model.

\paragraph{5.) Interactive Visualization Tool} Our interactive display shows the full computation of a two-layer additive risk model from beginning to end, without hiding any nonlinearities or computations from the user.\footnote{Anyone can interact with our tool at http://dukedatasciencefico.cs.duke.edu} Risk factors are colored according to their contributions to the model. The form of our model lends itself naturally to explaining predictions using risk factors, and understanding monotonicity constraints, through the visualization.
To create an explanation of an individual prediction, our interactive display first highlights the factors that contribute most heavily to the final prediction of the global model. Second, it shows patterns produced by \texttt{OptConsistentRule}.
Third, it provides case-based explanations. Our case-based reasoning method finds cases that are similar on important features to any current case that the user inputs.

Ultimately the value of a decision support tool lies with the value it provides to its users. Our model was evaluated by domain experts at the Fair Isaac Corporation (FICO), as part of the Explainable Machine Learning Challenge,\footnote{https://community.fico.com/s/explainable-machine-learning-challenge} where it received the FICO Recognition Award. This recognition is an important indicator that our framework is on the path to what financial decision-makers would find useful and appealing.
An overview of the visualization tool is provided in \ref{apx:vistool}.
A paper describing the competition entry appeared at the workshop
associated with the competition \citep{ChenEtAlFICO2018}.

The novel elements of the work are (i) the introduction of a decision support system for financial lending that includes several AI approaches that work together to form a holistic approach. This includes (ii) the globally interpretable two-layer additive risk model, which lends naturally to sparsity, decomposability, visualization, case-based reasoning, feature importance, and monotonicity constraints, (iii) the interactive visualization tool for the model and its local explanations, (iv) the use of the \texttt{OptConsistentRule} algorithm for high-support local conjunctive explanations, and (iv) the application to finance, indicating that black-box models may not be necessary in the case of credit-risk assessment.

\paragraph{Organization}
We provide background on the problem and formulate our solution in Section~\ref{sec:related}. We present our two-layer additive risk model in Section~\ref{gen_inst}, and evaluate it on the FICO dataset in Section~\ref{sec:evaluation} (additional experiments are reported in \ref{apx:german}).
In  Section~\ref{sec:visualization} we present our decision support system's interface and a mean to explain its predictions using risk factors.
In Section \ref{sec:summary_explanations}, we discuss consistent rule-based and case-based explanations of our global model. We review related work in Section~\ref{sec:literature review} and conclude in Section~\ref{sec:conclusion}.

\section{Problem Background and Problem Formulation}\label{sec:related}

Many companies are now providing explainable AI technologies for lending. This includes Equifax's proprietary NeuroDecision Technology \citep{equifax2018neurodecision}, and Additive Index Models described by Wells Fargo. 
What makes credit lending decisions different from other tasks in banking, such as direct marketing and financial trading, is the requirement of transparency. Such requirements are necessary because credit lending decisions are high-stakes decisions that affect whether someone could purchase a house or start a business. This is different than marketing or financial trading decisions where decisions can be made without explanation to the investors/users.

Despite the need for transparency in credit lending decisions, both Equifax's methodology and model are proprietary, and Wells-Fargo's model is also proprietary. Wells-Fargo's model (as described in their paper) cannot be used in practice for an additional reason: it does not seem to incorporate the monotonicity constraints that are required for lending, along with other types of constraints one might require for practical decision-making. 
Monotonicity and other constraints are essential to both interpretability and legal requirements for lending models \citep{fed2011guidance}; monotonicity constraints ensure that as risk factors increase, the estimated risk should also increase. However, typical machine learning methods for financial applications, such as decision trees and support vector machines for direct bank marketing \citep{moro2011data, moro2014data}, cannot be easily adapted to handle the monotonic relationship between certain features and predicted risk. The combination of monotonicity and transparency requirements are not naturally satisfied in machine learning approaches.

Traditionally, banks have constructed handcrafted models based on subscale scores to compute credit scores,
because such models are decomposable, interpretable, and obey monotonicity constraints: for example, FICO scores are computed from data grouped into five categories (i.e., subscales), payment history ($35\%$), amounts owed ($30\%$), length of credit history ($15\%$), new credit ($10\%$) and credit mix ($10\%$) (the weight of each subscale is included in parentheses) \citep{fico-score}.  Our evidence suggests that subscale models (like the one we designed) may not lose accuracy over more complex black box AI tools, if we design them flexibly and carefully -- this is a topic we explore when evaluating our global model.


\paragraph{Data} We work with the FICO dataset used in the Explainable Machine Learning Challenge \citep{ficocompetition}. Citing the data source: the data consists of an anonymized dataset of Home Equity Line of Credit (HELOC) applications made by real homeowners; the predictor variables come from anonymized credit bureau data; the target variable to predict is a binary variable that indicates whether a consumer was 90 days past due or worse at least once over a period of 24 months from when the credit account was opened. The dataset consists of approximately 10,000 observations and 23 features. For additional information about the description of each feature, see \citep{ficocompetition}. For a robustness check, we also work with the German Credit Dataset. The details of this dataset and the experimental results on this dataset are discussed in \ref{apx:german}.

\paragraph{Notation} We use $\{(\mathbf{x}_i, y_i)\}_{i=1}^N$ to denote the data, where $\mathbf{x}_i \in \mathbbm{R}^P$ is a vector of numerical features in the dataset. The labels are indicators of defaulting on a loan: $y_i \in \{0,1\}$. The records are assumed to be drawn i.i.d$.$ from a population that has been pre-screened to be at-risk for loan default, though the screening rule was not provided with the dataset. Let $\mathcal{P}$ represent the set of features with $|\mathcal{P}| = P$. The FICO dataset's $23$ features ($P = 23$) come with monotonicity constraints: credit risk must be decreasing with respect to certain features, and increasing with respect to others. Some features do not necessarily have monotonic relationships with credit risk. The German Credit Dataset consists of $P = 20$ features. 

\section{Two-Layer Additive Risk Model}
\label{gen_inst}
In this section, we describe our (global) two-layer additive risk model for credit risk prediction. We will use the FICO dataset for demonstration but a similar visualization can be created for other datasets (e.g., the German Credit Dataset discussed in \ref{apx:german}).

\subsection{Preprocessing}\label{sec:preprocessing}

We apply two preprocessing methods to ensure that our model satisfies monotonicity in some of the features, and can correctly handle missing values.

\noindent \textit{Monotonicity by piecewise constant design.}
We use a standard technique to ensure that our model respects the monotonic relationship between any given feature and the predicted risk. In particular, we transform the feature into \textit{one-sided} intervals and constrain the risk scores assigned to these one-sided intervals to be non-negative. More concretely, for a monotonically decreasing feature $x_{\cdot,p}$, we create the following binary features of one-sided intervals: $b_{p,1}(x_{\cdot,p})=\mathbbm{1}[x_{\cdot,p} < \theta_1],$ $b_{p,2}(x_{\cdot,p})=\mathbbm{1}[x_{\cdot,p} < \theta_2]$, ... ,  $b_{p,L_p}(x_{\cdot,p})=\mathbbm{1}[x_{\cdot,p} < \theta_{L_p}]$, and $b_{p,0}(x_{\cdot,p})=\mathbbm{1}_{[x_{\cdot,p} \text{ is not missing}]}$
where $b_{p,0}(x_{\cdot,p})$ is a special indicator for non-missing values,\footnote{We handle missing values by creating additional binary indicators to indicate that the values are missing.} $L_p$ is the number of binary features created for feature $p$ (not counting the special indicators for non-missing values and for missing values), and $\theta_1, \theta_2, ...$, and $\theta_{L_p}$ satisfy
$\theta_1 < \theta_2 < ... < \theta_{L_p}.$

To choose thresholds $\theta_l$ for creating the binary features, we followed a process that is somewhat similar to training a decision tree with a predetermined number of leaves on each feature. We worked with each feature separately, using its relationship to the outcome. Working with each feature separately is computationally efficient, allowed nonlinearities, and did not lead to a loss of accuracy compared to other methods. Specifically, to choose the thresholds for each feature, we used heuristic measures of ``purity'' of splitting the dataset by a specific candidate threshold on a feature, and then chose a threshold yielding the ``purest'' group. In particular, we used information gain, which is a commonly used metric in classification algorithms such as decision trees, to measure how informative splitting at a particular threshold is to predicting a target variable. We split the data repeatedly using a greedy approach, each time finding a threshold $\theta^*$ on a single feature that results in the largest information gain. Mathematically, this means that we set
$
\theta^* = \arg\min_{\theta} E(\theta),
$
where $E(\theta)$ is the entropy of splitting the subset of training data (that are currently being split) at $\theta$:
\begin{equation*}
E(\theta) = \frac{n_1}{n}\left(-\frac{n_1^+}{n_1}\log\frac{n_1^+}{n_1} - \frac{n_1^-}{n_1}\log\frac{n_1^-}{n_1}\right) 
+ \frac{n_2}{n}\left(-\frac{n_2^+}{n_2}\log\frac{n_2^+}{n_2} - \frac{n_2^-}{n_2}\log\frac{n_2^-}{n_2}\right).
\end{equation*}
In the equation above, $n$ is the number of training examples that are currently being split, $n_1$ (or $n_2$) is the number of training examples that are currently being split and that satisfy $x_{\cdot,p} < \theta$ (or $x_{\cdot,p} \geq \theta$, respectively), $n_1^+$ (or $n_1^-$) is the number of positive (or negative, respectively) training examples that are currently being split and that satisfy $x_{\cdot,p} < \theta$, and $n_2^+$ (or $n_2^-$) is the number of positive (or negative, respectively) training examples that are currently being split and that satisfy $x_{\cdot,p} \geq \theta$. In this way, our method flexibly decides where to ``bump'' the risk score for each individual feature.

\begin{figure*}[t]
        \centering
        \includegraphics[scale=0.4, trim={0 6.5cm 0 0}, clip]{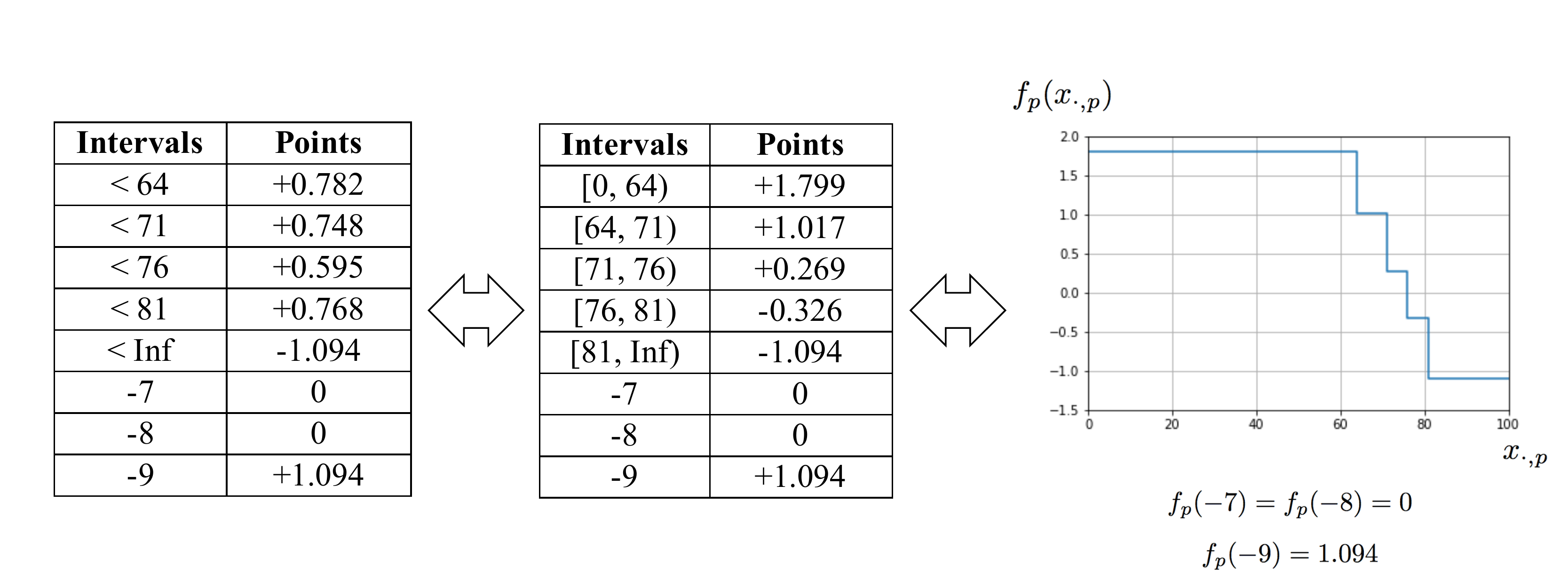}
        \caption{``External Risk Estimate'' feature. \textit{Left}: Risk scores $\beta_{p, l}$ assigned to one-sided intervals $x_{\cdot,p} < 64$, $x_{\cdot,p} < 71$, $x_{\cdot,p} < 76$, $x_{\cdot,p} < 81$, and $x_{\cdot,p} < \text{Inf}$. Note that $x_{\cdot,p} < \text{Inf}$ is equivalent to ``$x_{\cdot,p}$ is not missing,'' so the risk score $\beta_{p,0}$ assigned to it is allowed to be negative. $-7$, $-8$, and $-9$ are special values indicating that the feature value is missing, for three different reasons. The point values (e.g., +0.782, +.748, etc.) are learned using constrained logistic regression.  \textit{Middle}: Equivalent mathematical formulation of risk scores assigned to two-sided intervals $0 \leq x_{\cdot,p} < 64$, $64 \leq x_{\cdot,p} < 71$, $71 \leq x_{\cdot,p} < 76$, $76 \leq x_{\cdot,p} < 81$, and $81 \leq x_{\cdot,p} < \text{Inf}$. If a person has an external risk estimate of $x_{\cdot,p} = 33$, the person will satisfy the intervals $x_{\cdot,p} < 64$, $x_{\cdot,p} < 71$, $x_{\cdot,p} < 76$, $x_{\cdot,p} < 81$, and $x_{\cdot,p} < \text{Inf}$. This means that the person will receive $0.782 + 0.748 + 0.595 + 0.768 + (-1.094) = 1.799$ points. \textit{Right}: Plot showing the equivalent piecewise-constant risk-scoring function $f_p$ for the ``External Risk Estimate'' feature $x_{\cdot,p}$. Note that $f_p$ is indeed monotonically decreasing with respect to $x_{\cdot,p}$, for non-missing $x_{\cdot,p}$ values.}
        \label{fig:ExternalRiskEstimate}
    \end{figure*}

Note that all of the binary features $b_{p,1}(x_{\cdot,p}), b_{p,2}(x_{\cdot,p}), ..., b_{p,L_p}(x_{\cdot,p})$ created for the monotonically decreasing feature $x_{\cdot,p}$ use one-sided intervals. This choice was made because assigning risk scores to these one-sided intervals is equivalent to a weighted sum of the one-sided binary features, which yields a \textit{monotonic} piecewise constant risk-scoring function for the given feature if the risk scores for the one-sided intervals are non-negative. In other words, by enforcing constraints that the coefficients for $b_{p,1}(x_{\cdot,p}), b_{p,2}(x_{\cdot,p}), ...$, and $b_{p,L_p}(x_{\cdot,p})$ must be non-negative (the constraints are enforced during the training of the model; see Section~\ref{sec:training}), we shall guarantee a nonlinear monotonically decreasing relationship between the original numerical feature $x_{\cdot,p}$ and the predicted risk of default. To see this, let $f_p$ denote the risk-scoring function for the feature $x_{\cdot,p}$. Note that
\[
    f_p(x_{\cdot,p}) = \beta_{p,1} b_{p,1}(x_{\cdot,p}) + \beta_{p,2} b_{p,2}(x_{\cdot,p}) + ... + \beta_{p,L_p} b_{p,L_p}(x_{\cdot,p}) + \beta_{p,0} b_{p,0}(x_{\cdot,p})
\]
can be equivalently written as
\begin{eqnarray*}
    f_p(x_{\cdot,p}) &=& (\beta_{p,1}+\beta_{p,2}+...+\beta_{p,L_p}+\beta_{p,0})\mathbbm{1}[x_{\cdot,p}<\theta_1] \\
    &&+ (\beta_{p,2}+...+\beta_{p,L_p}+\beta_{p,0})\mathbbm{1}[\theta_1 \leq x_{\cdot,p} < \theta_2] + ... \\
    &&+ (\beta_{p,L_p}+\beta_{p,0})\mathbbm{1}[\theta_{L_p - 1} \leq x_{\cdot,p} < \theta_{L_p}] + \beta_{p,0}\mathbbm{1}[\theta_{L_p} \leq x_{\cdot,p}],
\end{eqnarray*}
which can be displayed as a traditional scoring system  \cite[e.g.,][]{mdcalc,ustun2016supersparse}. The risk-scoring function $f_p$ is  piecewise-constant, and can be visualized in a plot. In Figure~\ref{fig:ExternalRiskEstimate}, we show the transformation from risk scores for one-sided intervals to risk scores for two-sided intervals, and plot the piecewise-constant risk-scoring function, for the ``ExternalRiskEstimate'' feature, which is monotonically decreasing with respect to credit risk.

\textit{If coefficients $\beta_{p,1}, \beta_{p,2}, ...$, and $\beta_{p,L_p}$ are non-negative, it is easy to see that the risk-scoring function $f_p$ for the feature $x_{\cdot,p}$ is monotonically decreasing.} Note that we do not need to enforce the non-negativity constraint on $\beta_{p,0}$ to make $f_p$ monotonically decreasing.
If instead, we would like to constrain $f_p$ to be monotonically increasing, we reverse the above one-sided inequalities ``$<$'' into ``$\geq$''. Of course, if a feature $x_{\cdot,p}$ has no desired monotonicity, we discretize the feature into two-sided intervals, create binary features of the form $b_{p,l}(x_{\cdot,p})=\mathbbm{1}[\theta_{l-1} \leq x_{\cdot,p} < \theta_l]$, and drop the non-negativity constraints on the coefficients (i.e., the assigned risk scores) for the intervals during training.
    

\noindent\textit{Categorical and special values:} We create binary variables for each category for each categorical variable. For example, in the German Credit Dataset, the categorial variable ``Housing'' has three values: ``rent,'' ``own,'' and ``for free.'' We created three special binary features $b_{p,1}(x_{\cdot,p})=\mathbbm{1}[x_{\cdot,p} = \text{``rent''}]$, $b_{p,2}(x_{\cdot,p})=\mathbbm{1}[x_{\cdot,p} = \text{``own''}]$, and $b_{p,3}(x_{\cdot,p})=\mathbbm{1}[x_{\cdot,p} = \text{``for free''}]$, and set $f_p(x_{\cdot,p}) = \beta_{p,1}b_{p,1}(x_{\cdot,p}) + \beta_{p,2}b_{p,2}(x_{\cdot,p}) + \beta_{p,3}b_{p,3}(x_{\cdot,p})$ to handle the categorical variable ``Housing'' in this case. We can also handle structural missingness in a similar way. For instance,
the FICO dataset comes with three special values ($-7$, $-8$, and $-9$),\footnote{According to the documentation of the FICO dataset \citep{ficocompetition}, the special values and their interpretation are: -9 indicates  no bureau record or no investigation, -8 indicates no usable/valid trades or inquiries, and -7 indicates condition not met (e.g. no inquiries, no delinquencies).} to denote that a particular feature value is missing. For each feature $x_{\cdot,p}$, we also created three special binary features: $b_{p,-7}(x_{\cdot,p})=\mathbbm{1}[x_{\cdot,p} = -7]$, $b_{p,-8}(x_{\cdot,p})=\mathbbm{1}[x_{\cdot,p} = -8]$, and $b_{p,-9}(x_{\cdot,p})=\mathbbm{1}[x_{\cdot,p} = -9]$, and added the terms $\beta_{p,-7}b_{p,-7}(x_{\cdot,p}) + \beta_{p,-8}b_{p,-8}(x_{\cdot,p}) + \beta_{p,-9}b_{p,-9}(x_{\cdot,p})$ to $f_p(x_{\cdot,p})$, to handle the missing values.

\subsection{Model Form and Design}\label{sec:model_design}
The first layer of the two-layer additive risk model consists of $K$ subscales, each of which are additive models, and each additive model's components are piecewise constant functions of the original variables (we discuss how features are discretized to form  piecewise constant monotonic functions in Section~\ref{sec:preprocessing}). We used $K = 10$ for the two-layer additive risk model trained on the FICO dataset.

\noindent\textit{Subscales:}
Using domain knowledge obtained from the data description, we partitioned the features $\mathcal{P}$ into different subsets called subscales:
$
    \mathcal{P} = \cup_{k=1}^K\mathcal{P}^{[k]},
$
where $\mathcal{P}^{[k]}$ is a subset of features assigned to the $k$-th subscale. For the FICO dataset, we created 10 subscales, each containing one to four original features, related by their domain description. For example, the features ``MSinceOldestTradeOpen'' (Months Since Oldest Trade Open), ``MSinceMostRecentTradeOpen'' (Months Since Most Recent Trade Open), and ``AverageMInFile'' (Average Months in File) are all related to the length of one's credit history, and these features are grouped into one subscale called ``Trade Open Time.'' For each subscale, we build a mini-scoring model for predicting the probability of failure to repay a loan, using only the features designated for the subscale. Mathematically, each subscale model predicts a probability of default, denoted by $r^{[k]}$ for the $k$-th subscale, as follows:
\begin{equation}\label{eq:subscale}
    \begin{split}
        r^{[k]}(\mathbf{x}) &= \sigma\left(b_k + \sum_{p\in \mathcal{P}^{[k]}} f_p(x_{\cdot,p})\right) 
        = \sigma\left(b_k + \sum_{p\in \mathcal{P}^{[k]}} \sum_{l \in \{-7, -8, -9, 0, 1, ..., L_p\}} \beta_{p,l} b_{p,l}(x_{\cdot,p})\right),
    \end{split}
\end{equation}
where $\sigma(\cdot)$ represents the logistic sigmoid function $\sigma(x) = 1/(1+e^{-x})$, and $b_k$ is the bias term for the $k$-th subscale. Note that the equation above can be viewed as a scoring model that assigns $\beta_{p,l}$ points to the binary condition $b_{p,l}(x_{\cdot,p})$, for features $x_{\cdot,p}$ that are assigned to the $k$-th subscale. Alternatively, we can view the equation as a scoring model that assigns risk scores according to the risk-scoring functions $f_p$, for each original feature $x_{\cdot,p}$ belonging to the $k$-th subscale.

\noindent\textit{Second-layer model.} The subscale probabilities are combined using a weighted sum and nonlinearly transformed into a final probability, denoted by $r$, of failure to repay a loan:
\[
r(\mathbf{x}) = \sigma\left(b + \sum_{k=1}^K \alpha_k r^{[k]}(\mathbf{x})\right),
\]
where $b$ is the bias term and $\alpha_k$ is the coefficient for the probability predicted by the $k$-th subscale. The contribution of the $k$-th subscale to the final prediction can be easily observed by its weighted subscale probability $\alpha_k r^{[k]}(\mathbf{x})$.

We call the model a ``two-layer additive risk model'' because we view the subscale models as the first (hidden) layer that transforms an input into subscale probabilities, and view the transformation of subscale probabilities into a final predicted probability as the second layer. Conceptually, our two-layer additive risk model has a form of a small (two-layer) neural network. However, it differs from a regular neural network in multiple ways: unlike our model, a regular neural network does not partition features into meaningful subgroups; nor does it transform the original features into monotonic piecewise-constant risk-scoring functions. It also does not maintain models that are individually meaningful for subgroups of features; each of our subscale models are individually meaningful as ``mini-models'' using a subset of the features.

\subsection{Model Training}\label{sec:training}


The simplest way to train coefficients 
$\boldsymbol{\beta}^{[k]} = \{\beta_{p,l}\text{ for } p \in \mathcal{P}^{[k]}, l \in [0, L_p] \}$
is to treat each subscale $r^{[k]}$ as an independent logistic regression model with $\{y_i\}_{i=1}^N$ being the target variables, using regularization (e.g., $\ell_2$) to prevent overfitting, and non-negativity constraints on the coefficients to enforce monotonicity. Mathematically, we minimize the (regularized) logistic loss of the $k$-th subscale subject to non-negativity constraints, to train the $k$-th subscale:
\begin{equation*}
\min_{b_k, \boldsymbol{\beta}^{[k]}} \frac{1}{N}\sum_{i=1}^N L(r^{[k]}(\mathbf{x}_i), y_i) + \|\boldsymbol{\beta}^{[k]}\|_2^2 \;\;\;\;\textrm{ subject to}
\end{equation*}
\[
\beta_{p,l} \geq 0 \quad \text{ for all monotonic features } x_{\cdot,p} \text{ and } l \in \{1, ..., L_p\}.
\]
$L$ denotes the logistic loss function: $L(\hat{y},y)=-y\log\hat{y} - (1-y)\log(1-\hat{y})$. After we have trained all the subscales, we train the final model similarly, by minimizing the (regularized) logistic loss of the final model subject to non-negativity constraints:
\begin{equation*}
\min_{b, \alpha_1, ..., \alpha_K} \frac{1}{N}\sum_{i=1}^N L(r(\mathbf{x}_i), y_i) + \sum_{k=1}^K \alpha_k^2 \quad \text{ subject to } \alpha_k \geq 0 \text{ for all } k \in \{1, ..., K\}.
\end{equation*}
We enforce the non-negativity constraints on the $\alpha_k$'s, so that the final risk probability monotonically increases with each subscale probability. After we have trained the subscales and the final model separately, it is possible to fine-tune the subscales and the final model jointly, using (stochastic) gradient descent with a small learning rate ($10^{-5}$), so that the non-negativity constraints will not be violated).

\section{Evaluation}\label{sec:evaluation}

We next evaluate the performance of our two-layer additive model on the FICO dataset (for robustness, \ref{apx:german} describes the results of an additional experiment on the German Credit data). 
The FICO dataset appears to represent a subpopulation of individuals whose risk of credit default is particularly challenging. Approximately half of the population would default on a loan. 

\begin{table}[ht]
\caption{Comparison of our two-layer additive risk model with baseline models, in terms of test accuracy, AUC, and average precision, on the FICO dataset. We performed 10-fold stratified cross validation. The numbers were obtained by averaging the results over the 10 held-out test folds. The word ``binary'' in parentheses indicates that the model was trained and tested using binary features as input features. The word ``original'' in parentheses indicates that the model was trained and tested using the original features as input features. Values in italics are not statistically significantly different ($p$-value greater than $0.05$ under the paired sample $t$-test) from the best performance (which was attained by our two-layer additive risk model).}
\label{table:numerical_results}
\centering
\begin{small}
\begin{tabular}{|c|c|c|c|c|}
\hline
Classifier                       & Monotonicity? & Test accuracy     & AUC             & Average precision \\ \hline
Two-layer additive risk model    & \checkmark & $\mathbf{0.738} \pm 0.020$ & $\mathbf{0.806} \pm 0.025$ & $\mathbf{0.799} \pm 0.024$ \\ \hline
Single-layer additive risk model & \checkmark & $\mathit{0.736} \pm 0.022$ & $\mathit{0.805} \pm 0.024$ & $\mathit{0.799} \pm 0.024$ \\ \hline
Logistic regression (binary)     &  \xmark          & $\mathit{0.731} \pm 0.023$ & $\mathit{0.801} \pm 0.028$ & $\mathit{0.791} \pm 0.034$ \\ \hline
Logistic regression (original)  &  \xmark  & $0.699 \pm 0.044$ & $0.771 \pm 0.030$ & $\mathit{0.777} \pm 0.044$ \\ \hline
CART (binary)    &  \xmark              & $0.712 \pm 0.022$ & $0.771 \pm 0.026$ & $0.772 \pm 0.023$ \\ \hline
CART (original)   &  \xmark                           & $0.702 \pm 0.019$ & $0.757 \pm 0.036$ & $\mathit{0.766} \pm 0.061$ \\ \hline
Random forest (binary)    &  \xmark                   & $0.697 \pm 0.017$ & $0.757 \pm 0.017$ & $0.754 \pm 0.018$ \\ \hline
Random forest (original)   &  \xmark                  & $0.679 \pm 0.047$ & $0.744 \pm 0.037$ & $0.752 \pm 0.058$ \\ \hline
Boosted trees (binary)   &  \xmark                & $0.723 \pm 0.024$ & $0.789 \pm 0.028$ & $0.781 \pm 0.033$ \\ \hline
Boosted trees (original)      &       \xmark     & $0.694 \pm 0.044$ & $0.763 \pm 0.039$ & $0.760 \pm 0.054$ \\ \hline
XGBoost (binary)                 &       \xmark     & $\mathit{0.732} \pm 0.022$ & $\mathit{0.800} \pm 0.029$ & $\mathit{0.793} \pm 0.032$ \\ \hline
XGBoost (original)               &     \xmark       & $\mathit{0.709} \pm 0.047$ & $\mathit{0.782} \pm 0.042$ & $\mathit{0.789} \pm 0.048$ \\ \hline
SVM: linear (binary)             &     \xmark       & $0.720 \pm 0.029$ & $0.795 \pm 0.027$ & $\mathit{0.787} \pm 0.034$ \\ \hline
SVM: linear (original)           &      \xmark      & $0.700 \pm 0.046$ & $0.771 \pm 0.031$ & $\mathit{0.777} \pm 0.044$ \\ \hline
SVM: RBF (binary)                &      \xmark      & $0.727 \pm 0.023$ & $0.799 \pm 0.022$ & $\mathit{0.793} \pm 0.025$ \\ \hline
SVM: RBF (original)              &      \xmark      & $0.697 \pm 0.042$ & $0.760 \pm 0.039$ & $0.757 \pm 0.045$ \\ \hline
NN: 2 layer, sigmoid (binary)    &      \xmark      & $0.731 \pm 0.020$ & $\mathit{0.800} \pm 0.027$ & $\mathit{0.793} \pm 0.031$ \\ \hline
NN: 2 layer, sigmoid (original)  &      \xmark      & $0.684 \pm 0.042$ & $0.750 \pm 0.037$ & $\mathit{0.757} \pm 0.063$ \\ \hline
NN: 2 layer, ReLU (binary)       &      \xmark      & $0.717 \pm 0.020$ & $0.784 \pm 0.028$ & $0.774 \pm 0.035$ \\ \hline
NN: 2 layer, ReLU (original)     &      \xmark      & $0.687 \pm 0.041$ & $0.764 \pm 0.026$ & $\mathit{0.768} \pm 0.045$ \\ \hline
NN: 4 layer, sigmoid (binary)    &      \xmark      & $0.727 \pm 0.021$ & $0.796 \pm 0.027$ & $0.787 \pm 0.030$ \\ \hline
NN: 4 layer, sigmoid (original)  &       \xmark     & $0.687 \pm 0.041$ & $0.752 \pm 0.042$ & $\mathit{0.760} \pm 0.054$ \\ \hline
NN: 4 layer, ReLU (binary)       &      \xmark      & $0.710 \pm 0.020$ & $0.775 \pm 0.025$ & $0.764 \pm 0.027$ \\ \hline
NN: 4 layer, ReLU (original)     &      \xmark      & $0.694 \pm 0.044$ & $0.764 \pm 0.032$ & $\mathit{0.768} \pm 0.056$ \\ \hline
NN: 8 layer, sigmoid (binary)    &      \xmark      & $0.722 \pm 0.022$ & $0.792 \pm 0.026$ & $0.783 \pm 0.030$ \\ \hline
NN: 8 layer, sigmoid (original)  &       \xmark     & $0.674 \pm 0.043$ & $0.750 \pm 0.031$ & $\mathit{0.760} \pm 0.051$ \\ \hline
NN: 8 layer, ReLU (binary)       &       \xmark     & $0.705 \pm 0.027$ & $0.767 \pm 0.033$ & $0.750 \pm 0.039$ \\ \hline
NN: 8 layer, ReLU (original)     &      \xmark      & $0.699 \pm 0.044$ & $0.764 \pm 0.043$ & $\mathit{0.766} \pm 0.051$ \\ \hline
Dash et al. (2018) \citep{Sanjeeb2018}  & \checkmark & $0.711 \pm 0.023$ & $0.621 \pm 0.042$ & $0.735 \pm 0.051$ \\ \hline
\end{tabular}
\end{small}
\normalsize
\end{table}

\begin{figure}[ht]
\begin{subfigure}{.5\textwidth}
  \centering
  \includegraphics[width=\linewidth]{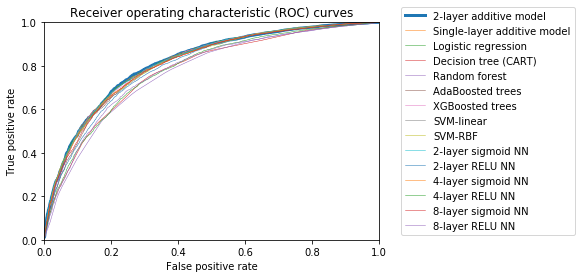}
  \caption{Receiver operating characteristic (ROC) curves of models using binary features.}
  \label{fig:roc}
\end{subfigure}%
\begin{subfigure}{.5\textwidth}
  \centering
  \includegraphics[width=\linewidth]{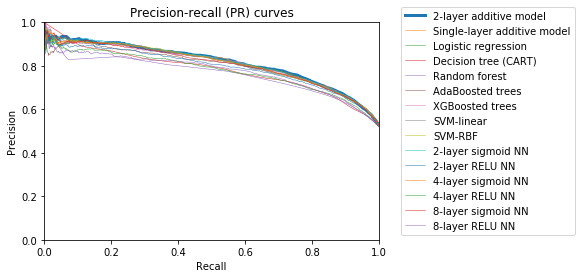}
  \caption{Precision-recall (PR) curves of models using binary features.}
  \label{fig:pr}
\end{subfigure}
\caption{ROC and PR curves of models using binary features.}
\label{fig:roc_pr}
\end{figure}

Table \ref{table:numerical_results} shows the test accuracy, the area under the receiver operating characteristic curve (AUC), and the average precision of our two-layer additive risk model, in comparison with other baseline models. The AUC is a measure of how well a classifier ranks credit risk of individuals. A model with a higher AUC gives a better credit-risk ranking of individuals than one with a lower AUC. The average precision is the area under the precision-recall curve, which is a measure of how well a classifier captures true high-risk individuals among individuals who are predicted to be of high risk. A model with a higher average precision has a lower false positive rate in general, and is less likely to misclassify a low-risk individual as a high-risk individual.  We performed 10-fold stratified cross validation, and used 9 folds for training and 1 fold for testing for 10 different training/testing splits. Table \ref{table:numerical_results} was obtained by averaging the results on the 10 held-out test folds -- both the mean and the standard deviation were computed for each performance metric over the 10 held-out test folds. The single-layer additive risk model is mathematically the same as a subscale model defined by Equation (\ref{eq:subscale}) with $\mathcal{P}^{[k]} = \mathcal{P}$, and is simply a large scoring model that assigns $\beta_{p,l}$ points to the binary condition $b_{p,l}(x_{\cdot,p})$ for all features $x_{\cdot,p}$ in $\mathcal{P}$. The single-layer additive risk model was trained using the same procedure (with the same non-negativity constraints) as a subscale model in the two-layer additive risk model, so it also respects the monotonic relationship between a monotonic original feature and the predicted risk of default. We performed a paired sample $t$-test between the performance results on the 10 test folds of our two-layer additive risk model and those of the baseline models, to test the statistical significance of the difference between the average performance value (averaged over 10 test folds) of our two-layer additive risk model, which attains the best performance, and the average performance values of the baselines. Values in italics in Table \ref{table:numerical_results} are not statistically significantly different (i.e., $p$-value greater than $0.05$) from the performance of our two-layer additive risk model. We make the following observations:

\textit{Binary features tend to outperform original features.} All models were trained and tested on the same training/testing splits of the original dataset (i.e., using original features), as well as the dataset where the original features have been transformed into binary features (as described in Section~\ref{sec:preprocessing}). The transformation of the original features into binary features is part of the two-layer additive risk model, as well as the single-layer additive risk model. As we can see in Table \ref{table:numerical_results}, all classifiers perform better using binary features as input features, compared to using original features as input features, for all three performance metrics we use. This shows that our discretization scheme captures the risk factors for default well.

\textit{All models using binary features tend to perform similarly.}
We also observe that all models that use binary features (including our two-layer additive risk model, as well as the single-layer additive risk model) perform similarly -- we plotted the average receiver operating characteristic (ROC) curves (Figure \ref{fig:roc}) and the average precision-recall (PR) curves (Figure \ref{fig:pr}) of all classifiers that use binary features, and observe that the ROC/PR curves of these classifiers are very close. The mean test accuracy, AUC, and average precision of our two-layer additive risk model are slightly higher than other baseline models (statistically significantly better than CART, random forest, boosted decision trees, and some neural networks), despite the fact that our model was more constrained than other baseline models. In particular, it is interesting to compare our two-layer (or single-layer) additive risk model with the logistic regression model that uses binary features as input. A major difference between our two-layer (or single-layer) additive risk model and the logistic regression model that uses binary features is that the former respects the monotonicity constraints while the latter may not. This shows that enforcing monotonicity constraints during training may not necessarily hurt the generalization performance of a classifier. It is also interesting to compare our two-layer additive risk model with the two-layer neural network (with sigmoid nonlinearity in its hidden layer) that uses binary features as input. These two models essentially have the same form, except that our two-layer additive model has very sparse connections in the first layer -- each binary input feature is connected to only one unit (i.e., one subscale) in the first layer. In addition, our two-layer additive model respects the monotonicity constraints but a two-layer neural network may not. Despite its simplicity and decomposability, and despite the enforcement of monotonicity constraints during training, the performance of our two-layer additive risk model is on par with, or better than, the two-layer neural network (and deeper neural networks). This shows that restricting the model to its interpretable form does not necessarily hurt performance. Lack of monotonicity in lending models may lead to undesirable model behavior -- for example, a person who has a longer credit history (e.g., as measured by the ``average months in file'' in the FICO dataset) could be predicted to have a higher risk than one with a shorter credit history. This could lead to ``counter-intuitive'' explanations for why a person would be denied a loan, and even violations of fair lending regulations.

\textit{Interpretability arises from a decomposition of the model into simpler subscale scores.}
With respect to interpretability, random forest, boosted decision trees, support vector machine (SVM) using RBF kernel, and fully-connected neural networks are too complex to be visualized. The single-layer additive risk model, logistic regression, and the linear SVM model can be visualized, but they lack the decomposability that the two-layer additive risk model has; the two-layer additive risk model has  ``mini'' models for each subscale, each of which produces a probability, and the subscale probabilities are combined in the second layer. A linear classifier does not calculate any of the subscale probabilities, and hence loses the subscale structure inherent in our two-layer additive risk model.

\section{Model Visualization and Explaining Predictions Using Risk Factors}\label{sec:visualization}

\begin{figure*}[t]
    \centering
    \includegraphics[width=.65\linewidth]{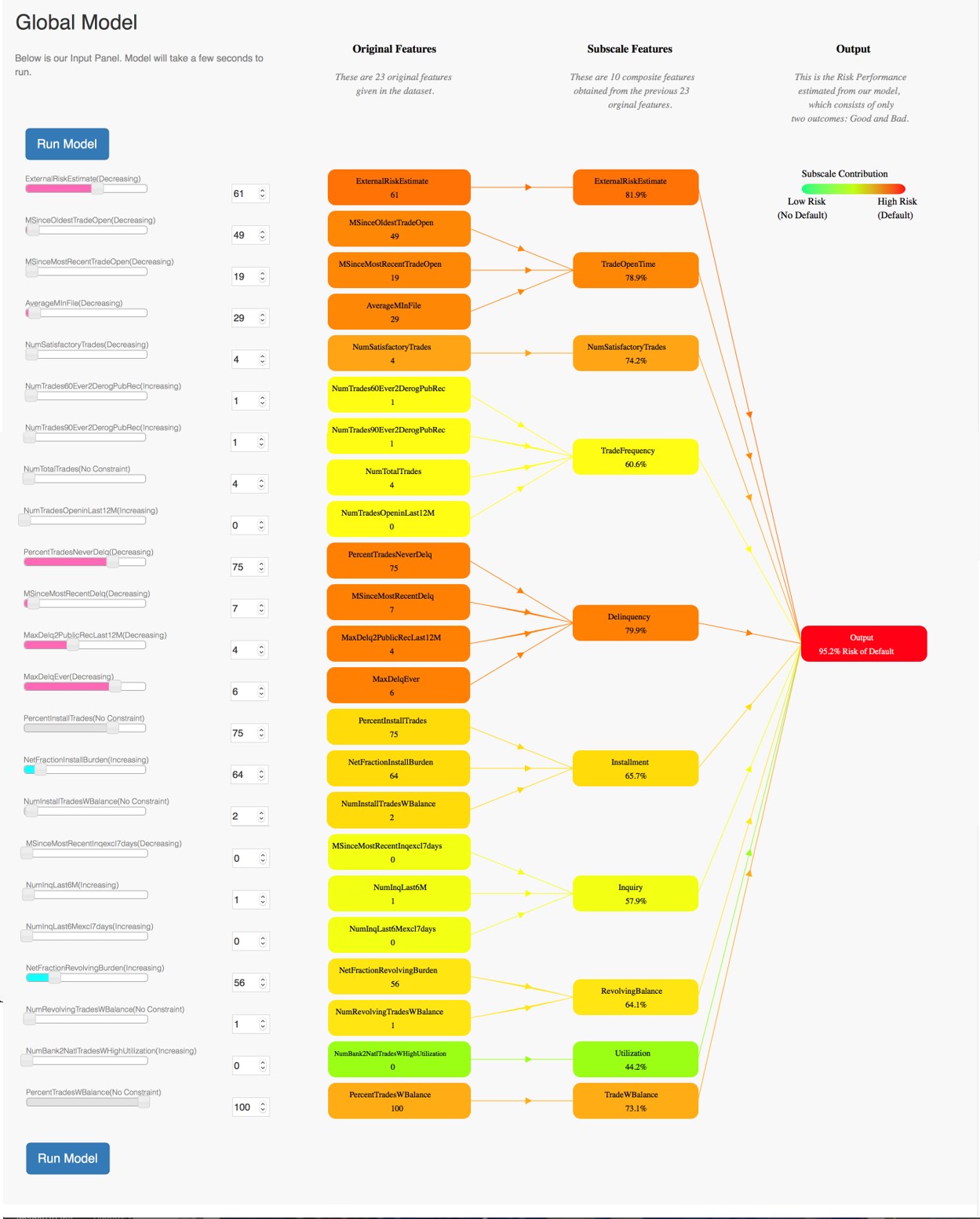}
    \includegraphics[width=.34\linewidth]{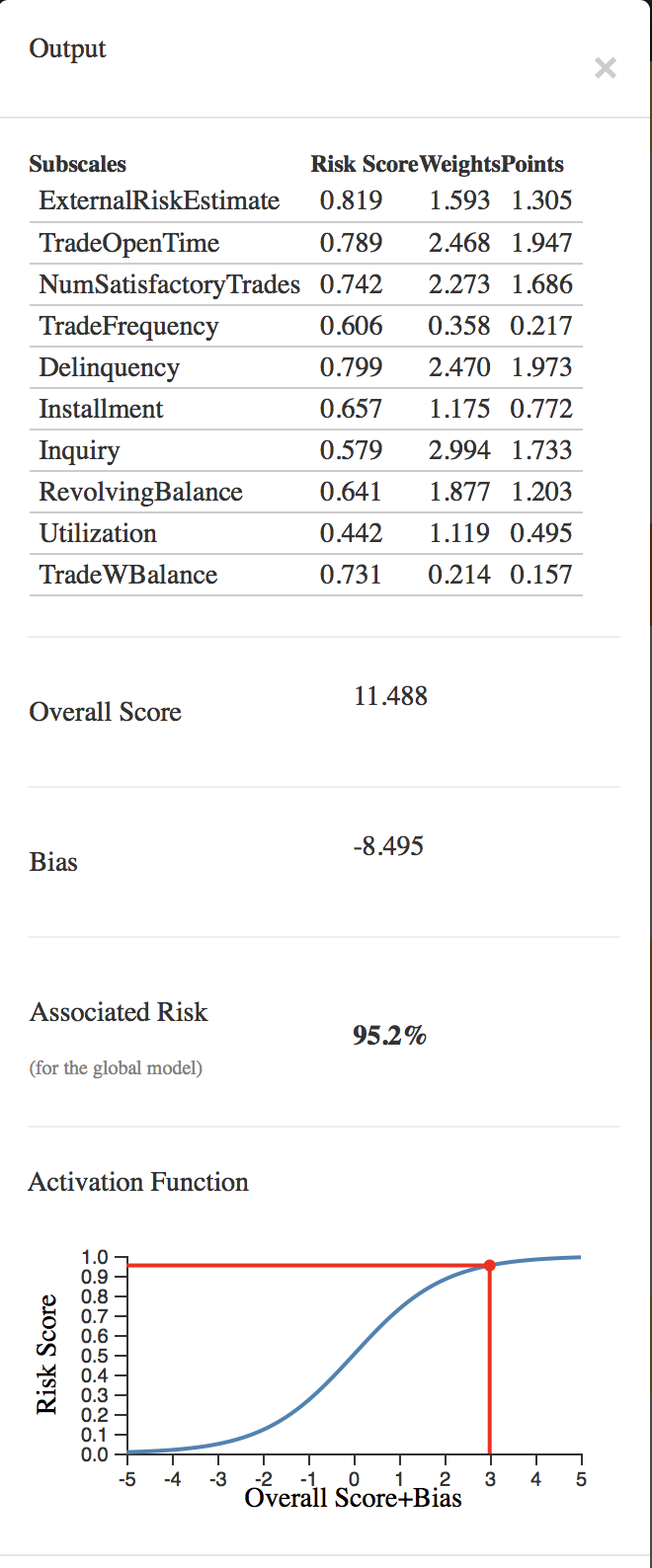}
    \caption{\textit{Left}: Snapshot of visualization tool showing the global model. Colors indicate contribution to the final score. The 23 feature values are entered on the left. \textit{Right}: Final combined score pop-up.}
    \label{fig:FullModel}
\end{figure*}

\begin{figure*}
        \centering
        \includegraphics[width=.3\linewidth]{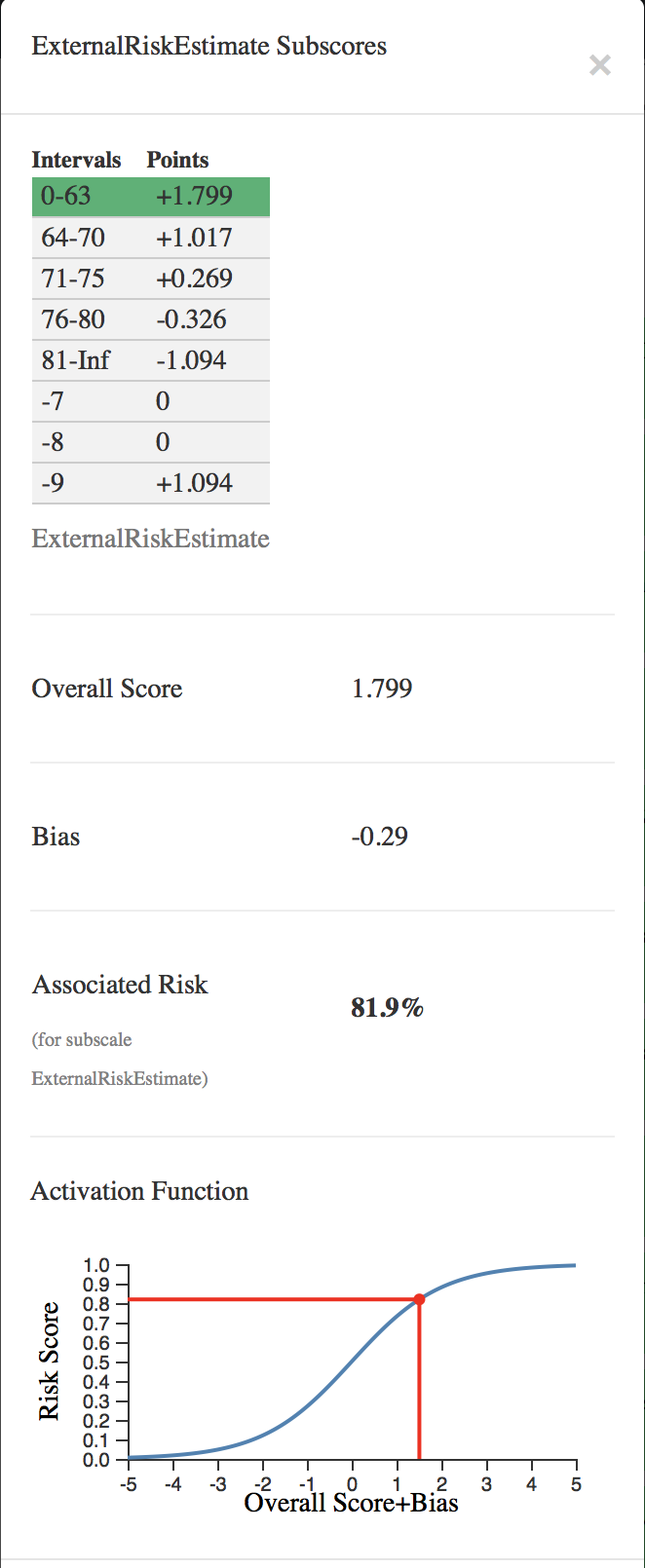}\;\;
         \includegraphics[width=.61\linewidth]{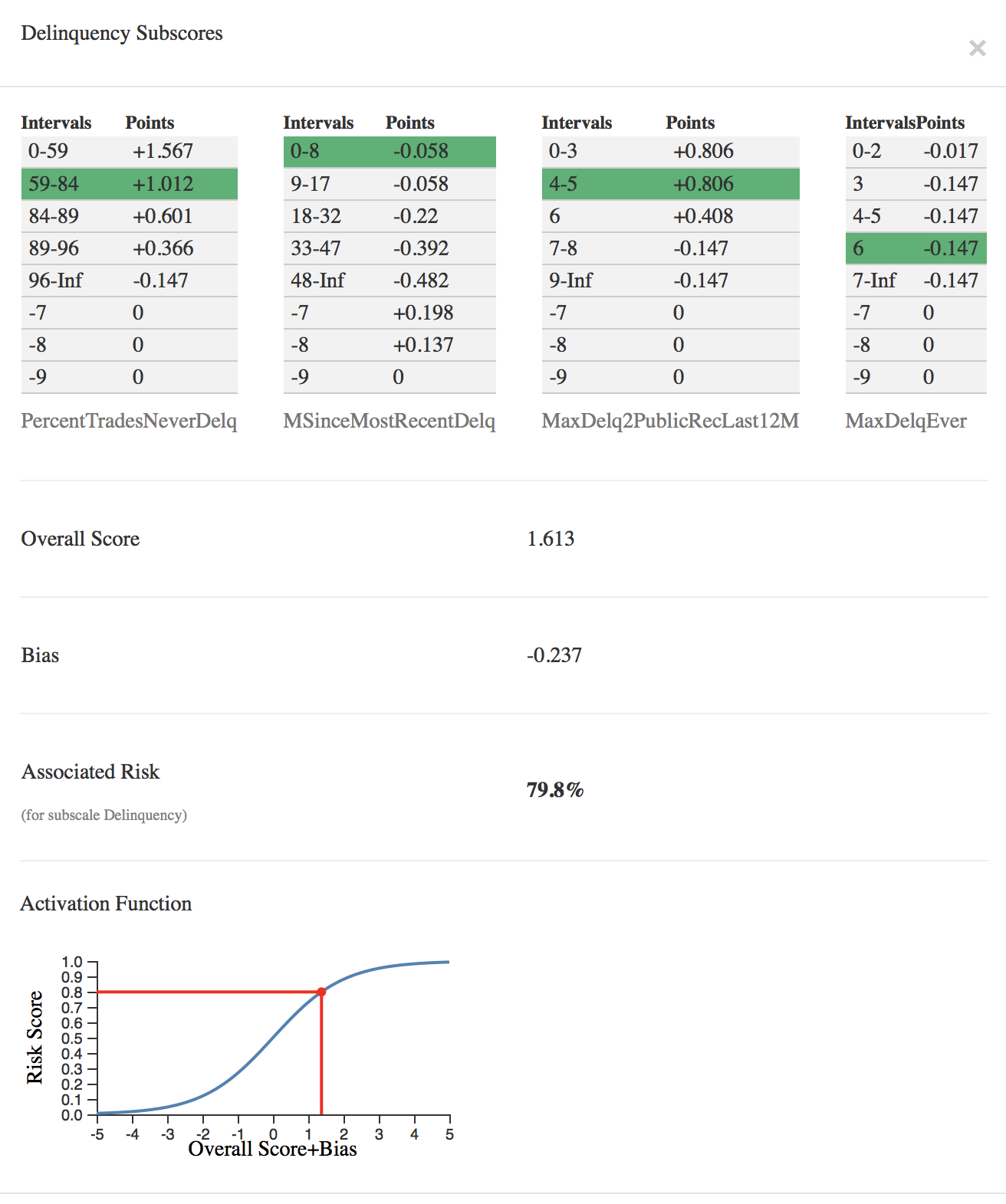}
        \caption{The ``External Risk Estimate'' and ``Delinquency'' Subscales.  \textit{Left}: The transformation of points into a risk estimate, which is 81.9\% for this person. The risk estimate for the subscale is meaningful as its own ``mini-model'' of risk, based only on the ExternalRiskEstimate feature, which represents a consolidated version of risk markers.  \textit{Right}: A different subscale, which uses multiple features associated with delinquency, including the percent of trades never delinquent, the months since most recent delinquency, maximum delinquency in the public record in the last 12 months, and the maximum delinquency ever recorded. See \citep{ficocompetition} for the description of the features.}
        \label{fig:subscales}
    \end{figure*}

A visualization of the full model is shown in Figure~\ref{fig:FullModel}. The subscale-predicted risks and the final predicted risk are color-coded, where red color indicates a higher predicted probability of default on a loan. 
We have developed an interactive visualization tool of our two-layer additive risk model, which is publically available at \url{http://dukedatasciencefico.cs.duke.edu}. In our interactive tool, the 23 input feature values can be entered on the left. Clicking on any of the ten subscales (in the middle colored layer) reveals a pop-up window with the calculation, as shown in Figure~\ref{fig:subscales} for two subscales. The final combination of subscales is shown in Figure~\ref{fig:FullModel} (right panel). An overview of the visualization tool is provided in \ref{apx:vistool}.

    %

\subsection{Explaining Predictions Using Risk Factors}\label{sec:risk_factors}

Our two-layer additive risk model comes naturally with a way to identify factors that contribute heavily to the final prediction.
Table~\ref{tab:ImportantFeatures} shows a list of four factors that are important for predicting an observation called ``Demo 1'' on our interactive visualization tool, to have $95.2\%$ risk of default (i.e., bad risk performance). 

\begin{table*}[th]
    \caption{An example of the most important contributing factors learned for observation ``Demo 1.'' MaxDelq2PublicRecLast12M is the maximum delinquency in the public record in the last 12 months, PercentTradesNeverDelq is the percent of trades never delinquent, and AverageMInFile is the average months in file.}
    \label{tab:ImportantFeatures}
    \centering
    \begin{tabular}{c|l}
\toprule
         & \multicolumn{1}{c}{\textbf{Most important contributing factors}} \\ \hline 
      1   & MaxDelq2PublicRecLast12M is 6 or less \\
          & (from the most important subscale, Delinquency)\\ 
      2 & PercentTradesNeverDelq is 95 or less \\
          & (from the most important subscale, Delinquency)\\ 
      3 & AverageMInFile is 48 or less \\
        & (from the second most important subscale, TradeOpenTime) \\ 
      4 & AverageMInFile is 69 or less \\
        & (from the second most important subscale, TradeOpenTime) \\
         \bottomrule
    \end{tabular}
\end{table*}
To identify the factors, we first identify the two most important subscales and then the most important factors within each subscale.
The importance of each subscale in the final model is determined by its weighted score, which is the product of the subscale's output and its coefficient -- the larger the product, the more it contributes to the final risk. 

For example, for ``Demo 1,'' the two most important subscales are Delinquency (with points of 1.973) and TradeOpenTime (with points of 1.947). Then, within each of the two subscales, we find two factors that contribute the most to that particular subscale's risk score. 
The two most important factors for each subscale are determined likewise by the product of the coefficient of each binary feature and the value of the binary feature itself. 
We finally output those binary features and their corresponding values as the most important contributing factors to the prediction made by our model.

The factors are grouped by subscales and are displayed in decreasing order of importance (within each important subscale) to the global model's predictions.

\section{Consistent Rule-based and Case-based Explanations}\label{sec:summary_explanations}

In addition to predicting risk using a globally interpretable model, we apply the algorithm  \texttt{OptConsistentRule} (which is our own practical adaptation of algorithms developed in \citep{ShaposhnikRu18}) to generate consistent rules that summarize broad patterns of the classifier with respect to the data. They do not explain the global model's computations; instead, they provide useful patterns; such patterns have been a popular form of explanation in the state-of-the-art literature on model explanations \citep{ribeiro2018anchors}. The patterns are globally consistent in the sense that \textit{all} observations satisfying the pattern's rule have the same predictions from the global model. The rule's prediction agrees with the global model consistently for all observations satisfying the rule.


As an example, consider Observation 6 in the FICO dataset, for which the global model predicts a high risk of default. Figure~\ref{fig:rule_example} shows   the rule-based sparse summary-explanation that includes Observation 6, which is returned by \texttt{OptConsistentRule}.
The rule explains that our global model predicts high-risk for Observation 6, because Observation 6 satisfies the conditions in the rule, and \textit{all} 700 previous cases that also satisfy the conditions in the rule are predicted as high-risk by our global model. A rule returned by \texttt{OptConsistentRule} is \textit{globally-consistent}, in the sense that there exists no previous case that satisfies the conditions in the rule but is predicted differently, by the global model, from what is stated in the rule. In contrast, explanations (from other methods) that are not consistent may hold for one customer but not for another, which could eventually jeopardize trust (e.g., ``That other person also satisfied the rule but he wasn't denied a loan, like I was!'')

\begin{figure}
    \centering
\begin{center}
\noindent\fbox{%
    \parbox{0.7\textwidth}{%
        For \textit{all} 700 (7.1\%) people where: 
        \begin{itemize}
            \item ExternalRiskEstimate$\leq 63$ , and
            \item  NetFractionRevolvingBurden$\geq 73$, 
        \end{itemize}
         the global model predicts a high risk of default.
    }%
}
\end{center}
\vspace*{-13pt}
    \caption{Example of a globally-consistent local explanation. It states that for all 700 people that have a poor risk estimate provided for them by an external source, and their net fraction of revolving burden (the revolving balance divided by credit limit) is more than 73\%, they have all been predicted to default by the global model.}
    \label{fig:rule_example}
\end{figure}

In what follows we introduce concepts from \citep{ShaposhnikRu18} for completeness. Specifically, we formalize the discussion on consistent rules in Section~\ref{sub:rules_notation}, and address aspects of optimization in Section~\ref{sub:rules_optimization}. In Section~\ref{sub:consistent-cases} we describe how to use consistent rules to identify similar cases for case-based explanations.

\label{headings}
\subsection{Notation and definitions}\label{sub:rules_notation}

Since we have transformed our original dataset into one with only binary features, it is now sufficient to consider a $P$-dimensional binary design matrix $\{\mathbf{x}_i\}_{i=1}^N$, that is, $x_{i,p}\in\{0,1\}$ for every $i$ and $p$. Let $h^{\m}:\{0,1\}^P\rightarrow \{0,1\}$ denote a classifier that was trained using the dataset, and let  $\{h^{\m}(\x_i)\}_{i=1}^N$ denote the vector of labels generated by the model $h^{\m}$ (the term ``global'' is meant to distinguish it from the ``local'' explanation that will be generated for each observation). 

Assume ${R}\subseteq \mathcal{P}$ is a subset of features and ${a}\in\{0,1\}$ is a label. The \textit{rule} ${R}\Rightarrow {a}$ describes the following binary function/classifier:
\[ h^{{R}\Rightarrow {a}}\left(\mathbf{x}\right) =
  \begin{cases}
    {a}       & \quad \text{if } \left(\prod_{p\in {R}} x_{\cdot,p}\right)=1 \\
    1-{a}     & \quad \text{otherwise.}
  \end{cases}
\]
That is, for a given observation $\mathbf{x}$, the rule ${R}\Rightarrow {a}$ predicts ${a}$ based on the projection of the observation onto the subspace of features ${R}$; specifically, by applying a logical AND operator on the subset of features in ${R}$. 

Let $\x_e$ and $h^{\m}(\x_e)$ respectively denote an observation and the corresponding model prediction that we wish to create a rule for. (Think of person $e$ as the loan applicant. We want to find a summary rule of  similar applicants who have the same prediction, which is $h^{\m}(\x_e)$.)
A rule ${R}\Rightarrow h^{\m}(\x_e)$ is said to be a globally-consistent rule-based summary-explanation for $\mathbf{x}_e, h^{\m}(\x_e)$ with respect to the model $h^{\m}$, the prediction $h^{\m}(\x_e)$, and the design matrix $\{\mathbf{x}_i\}_{i=1}^N$, if the rule's predictions, denoted as $h^{\e}(\x)$ for observation $\x$, satisfy the following conditions: 

\begin{enumerate}
	\item (Relevance) $h^{\e}(\textbf{x}_e)=h^{\m}(\textbf{x}_e)$, that is, $x_{e,p}=1$ for every $p\in {R}$. \;\;(The summary-explanation is valid for observation $e$.);
    \item (Global consistency) There is no observation $\textbf{x}_i$ for which  $h^{\m}(\x_i)=1-h^{\m}(\x_e)$ and 
    $h^{\e}(\x_i)=h^{\m}(\x_e)$.
    That is, if $i$ has the opposite global prediction as $e$, then $e$'s explanation must disagree with the global model's prediction of point $i$ (in which case observation $i$ must not obey the rule). 
\end{enumerate}
The second condition establishes the consistency of a summary-explanation. It requires any training observation satisfying the explanation's conditions ${R}$ to have the same label as our special observation $e$.
Hence, the summary-explanation ${R} \Rightarrow h^{\m}(\x_e)$ agrees with how the global model makes its predictions, and it is in this sense that our summary-explanation is consistent with the global model.

The quality of a rule ${R}\rightarrow h^{\m}(\x_e)$ is measured using two criteria:
\begin{itemize}
\item \textit{Sparsity} -- the cardinally of ${R}$, that is, $|{R}|$. This captures to a certain degree the level of interpretability of the rule, where more sparse rules are considered more interpretable. 
\item \textit{Support} -- the number of observations in the dataset that satisfy the rule, namely $|\{\mathbf{x}_i:\left(\prod_{p\in {R}} x_{i,p}\right)=1\}|$. This serves as a measure of coverage for the applicability of the rule, where rules with larger support are considered more trustworthy by users.
\end{itemize}

To ensure that the rules also capture predicates of the form $\x_{\cdot,p}=0$ (and not only 
$\x_{\cdot,p}=1$), we expand the feature space by adding the complementary feature $\x^C_{\cdot,p}=1-\x_{\cdot,p}$ for each feature   $\x_{\cdot,p}$, so that a rule that contains the predicate $\x^C_{\cdot,p}=1$ is equivalent to including instead the predicate $\x_{\cdot,p}=0$.


\subsection{Optimization}\label{sub:rules_optimization}


To find globally-consistent rules for binary datasets,  \cite{ShaposhnikRu18} formulate two optimization problems and solves them using mixed-integer programs: 
\begin{enumerate}
    \item The algorithm \texttt{BinMinSetCover} computes a globally-consistent rule with optimal sparsity by solving the following optimization problem:
\begin{equation}\label{eq:min_cardinality1}
\begin{array}{ll@{}ll}
\min_{\{I_p\}_{p\in \mathcal{P}}} &  \sum_{p\in \mathcal{P}} I_p      &\textrm{ (optimize sparsity)}\\
\text{s.t.}& \sum_{p\in \mathcal{P}} I_p \cdot(\mathbbm{1}[x_{e,p} = 1] - 1) \geq 0  &   \textrm{ (summary-explanation is relevant)}\\
\qquad\;\; \forall i\in [N]: & 
\textrm{If } h^{\m}(\textbf{x}_i)=1-h^{\m}(\textbf{x}_e) &   \textrm{ (summary-explanation is consistent)}\\
& \textrm{Then } \sum_{p\in \mathcal{P}} I_p \cdot\mathbbm{1}[x_{i,p} = 0] \geq 1,   & 
\end{array}
\end{equation}
     where $I_p$ is a decision variable indicating that the feature $p$ is part of our rule. Note that all other terms in the formulation are constant. This formulation would be solved using an integer programming solver.
    
    \item The algorithm \texttt{BinMaxSupport} computes a globally-consistent rule that balances support (the number of points satisfying the rule) and the number of terms comprising the rule (i.e., sparsity). The objective in this optimization problem is: 
    $$\max_{\{r_i\}_{i\in [N]}, \{I_p\}_{p\in \mathcal{P}}}\qquad w_s\cdot \sum_{i\in [N]} r_i -  w_c\cdot\sum_{p\in \mathcal{P}} I_p, \;\;\textrm{ (maximize support and minimize complexity)}$$ 
    where $r_i$ is a binary variable indicating that observation $i$ satisfies the rule (and therefore $\sum_{i\in [N]} r_i$ counts the support). The constraints includes those of Equation~\eqref{eq:min_cardinality1} (that is, relevance and global consistency), in addition to a maximal sparsity constraint (that is, $ \sum_{p\in \mathcal{P}} I_p\leq \textrm{MAX\_SPARSITY}$), and constraints that ensure $r_i$ correctly indicates that a rule is satisfied by observation $i$ (that is, $\sum_{p\in \mathcal{P}} I_p\cdot \mathbbm{1}[x_{i,p} = 0] \leq M\cdot (1-r_i)$, where $M$ is a large constant used in the Big M method).
\end{enumerate}
To generate summary-explanations in our visualization tool, we introduce our own adaptation of the algorithms called \texttt{OptConsistentRule}, which is designed to provide a practical solution by solving a sequence of optimization problem instances of types \texttt{BinMinSetCover} and \texttt{BinMaxSupport}.
Roughly speaking, \texttt{OptConsistentRule} first optimizes for sparsity by solving \texttt{BinMinSetCover} to find OPT\_SPARSITY --  the smallest number of features for which there exists a globally-consistent rule. We then use this solution as an initial solution to \texttt{BinMaxSupport} which maximizes support subject to the constraint that there are at most OPT\_SPARSITY features. This provides us with the rule whose support is maximal among all rules whose sparsity is optimal.
We then resolve \texttt{BinMaxSupport} two more times, each time increasing MAX\_SPARSITY by one and reusing previously obtained solutions as initial solutions for the current optimization problem. This is done to relax the maximal sparsity constraint in order to improve the support (at the cost of slightly worsening sparsity). 
\texttt{OptConsistentRule} returns the most sparse rule whose support exceeds a user defined threshold.
We discuss additional implementation details in \ref{apx:implementation_details}.

\subsubsection{Additional examples of rule-based explanations}

Figure~\ref{fig:additional_rules} presents additional examples of globally-consistent summary-explanations generated from the data and our two-layer model. 
We see that our approach can be used to explain both high- and low-risk predictions: the first two explain high-risk predictions while the third explains a low-risk prediction. We observe that the explanations agree with the monotonicity constraints. For example, the risk is non-increasing in all three features used in the first summary-explanation, and indeed the predicates indicate that these features are not large enough in the explained observation. Similarly, for the second observation the feature MSinceMostRecentDelq decreases risk, the feature NetFractionRevolvingBurden increases risk, and the risk is lower when NetFractionRevolvingBurden is missing (the inequalities indeed match this intuition). 
Note that these rules, which agree with the monotonicity constraints, were extracted without any knowledge about the global model except for its predictions on the data (that is, the algorithm is model agnostic).

\begin{figure}[ht]
    \centering
    \begin{center}
    \noindent\fbox{%
        \parbox{0.7\textwidth}{%
            For \textit{all} 812 (8.2\%) people where: 
            \begin{itemize}
                \item \textcolor{red}{ExternalRiskEstimate}$\leq 71$, 
                \item \textcolor{red}{MSinceMostRecentInqexcl7days} $\leq 1$, and
                \item \textcolor{red}{PercentTradesNeverDelq} $\leq 84$, 
            \end{itemize}
             the global model predicts a \textbf{high} risk of default.
        }%
    }    
    \noindent\fbox{%
        \parbox{0.7\textwidth}{%
            For \textit{all} 400 (4.1\%) people where: 
            \begin{itemize}
                \item \textcolor{red}{MSinceMostRecentDelq} $\leq 33$,
                \item \textcolor{blue}{NetFractionRevolvingBurden} $\geq 72$, and
                \item \textcolor{blue}{NetFractionInstallBurden is not missing ($\neq 8$)}
            \end{itemize}
             the global model predicts a \textbf{high} risk of default.
        }%
    }
    \noindent\fbox{%
        \parbox{0.7\textwidth}{%
            For \textit{all} 294 (3\%) people where: 
            \begin{itemize}
                \item \textcolor{red}{MSinceMostRecentInqexcl7days is missing ($-8$)}, and
                \item \textcolor{red}{ExternalRiskEstimate}  $\geq 81$, 
            \end{itemize}
             the global model predicts a \textbf{low} risk of default.
        }%
    }
    \end{center}
\vspace*{-13pt}
    \caption{Additional examples of globally-consistent local explanations. Explanations can be generated for both low- and high-risk predictions. The explanations agree with the monotonicity constraints (colored in \textcolor{red}{red} and \textcolor{blue}{blue} are features that decrease and increase risk, respectively). 
    From \citep{ficocompetition}, the meaning of the features is as follows. 
    ExternalRiskEstimate: consolidated version of risk markers, 
    MSinceMostRecentInqexcl7days:	months since most recent inquiry excluding the last 7 days, 
    PercentTradesNeverDelq: percent trades never delinquent,
    MSinceMostRecentDelq:	months since most recent delinquency, 
    NetFractionRevolvingBurden: net fraction of revolving burden (the revolving balance  divided by credit limit), 
    NetFractionInstallBurden: net fraction installment burden (the installment balance divided by original loan amount).}
    \label{fig:additional_rules}
\end{figure}

\subsubsection{Evaluation of summary-explanation}

We briefly note that \citep{ShaposhnikRu18} conducted a computational study on the FICO dataset where sparse explanations were generated for each of the 10,000 observations, based on all other observations. The running time consistently took less than 7 seconds, and the average sparsity was under three features. In addition, they generated summary-explanations for all observations in the FICO dataset. The average number of features in each explanation did not change (comparing with the optimal sparsity) and was equal to 2.9 when MAX\_SPARSITY was set to its smallest possible value; sparsity slightly worsened to 3.6 and 4.4 when MAX\_SPARSITY was increased by 1 and 2, respectively. They also found that the number of summary-explanations for which the support is less than 10 was 9.7\% of all observations and was equal to 4.7\%, 1.2\%, and 0.2\% of all observations for the maximal support solution when  MAX\_SPARSITY was increased by 0, 1, and 2, respectively. Clearly, this indicates a trade-off between support and sparsity; when the constraint on MAX\_SPARSITY is relaxed, the sparsity of the rule worsens and the rule becomes less interpretable; however, at the same time, the support of the rule increases, which improves the confidence in the resulting rule. Overall, even with an increase of up to 2 features, the average sparsity remains low, and the support is reasonably high.

\begin{figure*}[ht]
 \includegraphics[width=1.05\linewidth]{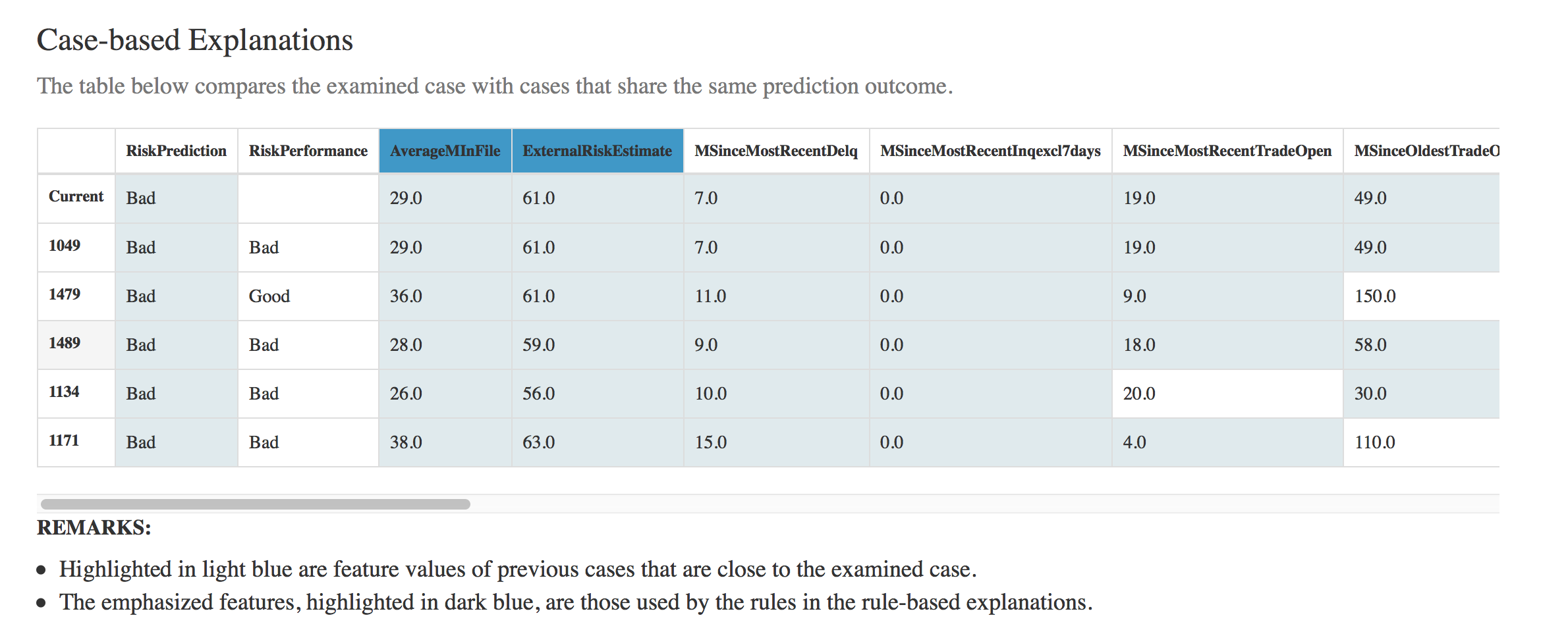}
  \caption{Case-based explanations. The current case is on the top line. Features description from \citep{ficocompetition} -- Risk Prediction: our model prediction, 
  RiskPerformance: the label (was consumer late on loan payments), 
  AverageMInFile: average months in file, 
  ExternalRiskEstimate: consolidated version of risk markers, 
  MSinceMostRecentDelq:	months since most recent delinquency, 
  MSinceMostRecentInqexcl7days:	months since most recent inquiry excluding the last 7 days, 
  MSinceMostRecentTradeOpen: months since the most recent trade open,   MSinceOldestTradeOpen: months since oldest trade open.}
  \label{fig:CaseBased}
\end{figure*}


\subsection{Case-based explanations}\label{sub:consistent-cases}

We propose a way to generate case-based explanations for our global model. 
Given a (previously unseen) test case, we identify similar cases from the FICO dataset to allow the user to manually validate the prediction made by our global model for the test case. To do this, we find all the cases in the dataset that satisfy the summary-explanation generated for the test case. We then rank these cases according to how many binary features (Section~\ref{gen_inst} discusses the discretization procedure) they share with the test case and present the five highest-ranked similar cases to the user. As an example, Figure~\ref{fig:ExternalRiskEstimate} illustrates the discretization of the feature ``ExternalRiskEstimate.'' If the features of two observations reside within the same interval they will be considered similar (e.g., ExternalRiskEstimate=73 and ExternalRiskEstimate=75 are considered similar). The total number of similar features is the similarity metric used to rank observations.
Figure~\ref{fig:CaseBased} gives an example of a case-based explanation: our visualization contains a table showing the current case (top row) and previous cases that are most closely related to the current case (all other rows). 




\section{Related Work}\label{sec:literature review}

\paragraph{Background on the Explainable Machine Learning Challenge}
There are a number of recent academic works in explainable credit risk modeling stemming from the 2018 Explainable Machine Learning Challenge
\citep{Sanjeeb2018, Rory2018, Steffen2018}, which, to our knowledge, was the first public challenge on explainable AI (XAI).  A notable solution came from \citet{Sanjeeb2018}, who developed a column generation framework and used it to build Boolean rules (in disjunctive normal form) for credit risk predictions. 
While sparse rules are interpretable in many domains, they have some serious drawbacks in loan decisions: (1) For loan decisions, risk calculations typically take into account several different aspects of the user's credit history, including delinquency, trade open times, and trade frequency. None of these are considered in the sparse rules of \citet{Sanjeeb2018}. Leaving these out might lead to over-simplistic risk models, and users might be concerned if the vast majority of their credit history statistics are ignored in their loan application. \citet{Elomaa94indefense} and \cite{Freitas:2014ic} stress that ``Humans by nature are mentally opposed to too simplistic representations of complex relations.'' (2) For loan decisions, real-valued risk predictions would generally be more relevant than a yes/no rule, since loan officers would typically consider a risk score and a probability of default, rather than just a yes/no from an algorithm. The lack of real-valued risk predictions in the sparse rules of \citet{Sanjeeb2018} means that the model is unlikely to rank the loan applicants well, in terms of their credit risk. This is reflected in the lower AUC score of the model of \citet{Sanjeeb2018}: while our two-layer additive risk model has an AUC around 0.80, and other baselines have AUCs above 0.74, the model of \citet{Sanjeeb2018} has an AUC around 0.62 (see Table \ref{table:numerical_results}). 


We note that the vast majority of entries in the competition did create black box models and explain them, despite the fact that interpretable models (like the one we created) were possible for this dataset.

\textit{Background on interpretability in decision support systems.}
Interpretability has been increasingly sought after in decision-support systems generally. For example, \citet{de2021spline} propose rule-based classifiers for interpretable customer churn prediction. \citet{kraus2019forecasting} use an interpretable deep learning approach to forecast remaining useful life of machinery, infrastructure, or other equipment. It is possible that the framework developed in our work could be used in these (or other) applications: a two-layer additive risk model that organizes all relevant features into a decomposable and visualizable model, respects monotonic relationships given with the dataset, and produces a numerical risk of default, can be extremely helpful in risk assessment for many domains. The visualization and summary-explanation techniques we propose are also generally useful.

\textit{Background on explanations of black boxes.}
One thing notably absent from this paper is a black box model that is ``explained.'' Explained black boxes
constitute the main thrust of the field of explainable machine learning. There are many reasons why one would not want to explain black box models \cite{Rudin19,RudinRa2019}, the most convincing reason being that we did not lose accuracy to construct a globally interpretable model instead of a black box. There have been several works within the past few years that showed why posthoc explanation methods such as LIME \citep{ribeiro2016should,ZhangSoSuTaUd2019} do not necessarily represent the underlying black box model, and how posthoc explanations are flawed in various other ways \citep{AdebayoEtAl18,SlackEtAl20}.  Explanations of black box models have been misinterpreted in the past, in some cases leading to influential scandals \citep[e.g., see][]{RudinWaCo2020}. Even if an explanation gives the same prediction as the global model, it could be inconsistent with the global model's actual calculations  \cite[see also][]{guidotti2018survey}. 
As discussed, an underlying assumption in the use of explanation tools is that one cannot construct a globally interpretable model that is as accurate as a black box. However, in the case of the FICO data, as we showed, this assumption is false.

There are several works that focus on counterfactual explanations for black box credit risk models \citep[e.g.,][]{Rory2018}; the work of \citet{Steffen2018} created an interactive visual analytics tools for such explanations. Counterfactual explanations could be built to assist with our globally interpretable model as well, but counterfactual explanations come with a challenge: one must elicit the cost from each user to alter a particular feature, which is perhaps much more difficult to obtain than the prediction itself. Sensitivity analysis \citep[e.g., ][]{cortez2013using} would also be relevant in understanding the behavior of models. Our summary-explanations are designed to show regions around a point where the model's predictions are not sensitive to changes in covariates.

\textit{Background on regulations.}
Even with regulations, such as the General Data Protection Regulation \citep{regulation2016general} and Equal Credit Opportunities Act \cite{cfpb}, that stipulates clients' right to an explanation, there is still little incentive for credit-rating companies to provide anything more than a modestly local explanation, even if a globally interpretable model existed.

\textit{Background on machine learning approaches for additive models and scoring systems.}
Additive scoring models (like the one we created here) are natural for real-valued features and can easily preserve monotonicity. Generalized additive models can be created using boosting methods (or any linear modeling methods), but these are generally not sparse \citep[e.g., see ][]{Caruana13} like our subscale models, which have only a few ``steps'' per feature. 

There are recent machine learning techniques to create medical scoring systems \citep{ustun2016supersparse, UstunRu2017KDD, UstunRu19, Sok2017, Sok12018, Sok22018, CarrizosaEtAl16} or scoring models for risk management \citep{escudero2015sdp}, which are sparse (unlike the additive modeling techniques discussed above). These techniques could potentially be used in the future to construct each of the subscales.

\textit{Goodhart's Law.} 
A possible objection to using transparent credit risk models is that an individual who is not credit-worthy may try to game the models to improve their credit score, leading to a situation described in Goodhart's law \citep{goodhart1984problems}. However, since credit scores should be aligned inversely with credit risk, actions taken to increase credit scores should actually lower credit risk; if this is not the case, the scoring system is not designed correctly: using inappropriate features, violating monotonicity, etc. No one should be able to have bad credit and a good credit score simultaneously if the credit score is properly constructed.


\section{Conclusion and Discussion}\label{sec:conclusion}

In this paper, we presented a holistic approach for designing a decision support system for financial lending that is both interpretable and accurate. Our approach includes: (1) a globally transparent two-layer additive risk model, which is a hybrid of subscale scoring models and a neural network, designed for model flexibility and accuracy; (2) a ranking mechanism to find important factors; (3) globally-consistent summary explanations to explain model behavior; (4) case-based explanations to show similar cases that are predicted in the same way as the unknown test case; (5) a decision support system that interactively visualizes of the entire model. We will offer some concluding remarks in this section.

\textbf{Lending Needs Transparency:} Loan decisions are high-stakes decisions. They affect large numbers of people, and whether they succeed in starting a business, in purchasing a house, or in other essential aspects that fundamentally affect people's lives. In the current era, where shocks in the economy have happened and challenges from aftershocks are likely to remain (due to pandemics, politics, and other events), transparency is paramount in models that make lending decisions on people.
One of the most important reasons for transparency is \textit{domain shift} (also called domain adaptation). Domain shift is where data from the past no longer represent the present. For instance, the number of recent past defaults on a loan have a different meaning, depending on when these defaults occurred with respect to an economic shock and what industry the applicant works in. Domain shift can compromise statistical models, in which case, these models need to be adjusted by the lender. Such adjustments are not easy to make unless one knows exactly what contribution each feature makes to the model. 

There are many other reasons that transparency is paramount, as we discussed earlier, ranging from checking for typographical errors to understanding variable importance, to providing reasons that are truly correct rather than approximately correct, to being able to understand exactly -- not approximately -- why one was denied a loan.

Our framework aims to illuminate how transparency can be achieved through several different mechanisms that work together to provide exact explanations for loan decisions.

\textbf{Competition Judging and Lessons Learned:} An important topic in interpretable machine learning, as well as in explainable AI, is to be able to judge the quality of an explanation or interpretable model. 
Since there is no single definition of interpretability, or explainability of black boxes, (nor should there be, see \cite[][]{Rudin19}) the question of how the competition was judged is relevant to how we might assess--and thus design--interpretable or explainable AI systems in the future.

Since the judging procedure was not announced in advance, none of the entrants knew prior to submitting their entries what the judging criteria would be. There were two evaluations: one created by the competition organizers, and an evaluation by FICO itself. 
As it turned out, in the first evaluation by the competition organizers, the competition's judging procedure for the challenge focused essentially on the \textit{ease of calculating} the explanation; judges were provided with partial explanations for only four observations, and asked to predict the black-box model's predictions for these four observations, for both global and local predictions. Judges were not permitted to use any of the visualization tools submitted by any of the entrants. 

Generally, visualization tools are particularly useful for \textit{counterfactual reasoning}: users of the visualization tools can determine in real time what happens to the predictions when a feature changes by adjusting it directly. Since the judges were not able to use any of the visualization tools, they were not able to reason this way.

More importantly, the judges were not able to evaluate the \textit{fidelity} of any of the local models in representing their global model, for any of the competition entries. This means that local models that were poorly fitted to the global models may have received high scores. (Also, it is not clear how our entry was provided to the judges since all predictions -- from both our globally interpretable model and the summary-explanations -- are both local and global predictions for every observation. If our rule-based explanations were provided as both local and global predictions, then they may have been easier to calculate than the global model.)

Unfortunately, this meant that the judging team was not able to judge some of the main criteria for explainability: fidelity of the explanation, or counterfactual reasoning. 

Since the competition's instructions were to produce a global black box and an explanatory local model, entries unfortunately did not have the opportunity to be assessed on the interpretability of their global model.

The second evaluation of the competition was performed by FICO itself, and not by the competition organizers. We do not have details on how the entries were judged. Our entry was ranked first by FICO's criteria, earning the \textit{FICO Recognition Prize} for the competition. This allowed us to understand that FICO actually did value a globally interpretable model above black-box models when given a choice between them.

There are some clear lessons learned from this competition. First, \textit{it is too often assumed that black-box models are necessary when they are not}. In many cases, black-box models are the mechanism by which a company profits -- this fuels what has been called ``the colossal asymmetry between societal cost and private gain in the rollout of automated systems'' \citep{Powles2018}. The case of credit scoring is a key example -- if the dataset from FICO is at all representative, our work demonstrates that \textit{perhaps we do not need black-box or proprietary models for credit scoring}.

A second lesson learned is that explainability should not always be judged according to ease of computation. Additive models are known to be interpretable/explainable but are not easy to compute by hand. It is well-known that sparsity, while useful for some applications, is not the only measure of interpretability \citep{Freitas:2014ic}. Scoring models may be more useful, but are more difficult to calculate, than models that predict only yes/no. Further, aspects such as fidelity of explanation, case-based reasoning, counterfactual reasoning and recourse, and the ability of the user to understand how variables are combined \textit{jointly} to form predictions are often more important than only ease of calculation of a local explanation (which was the judging criteria).

\textbf{Takeaways:} 
The main takeaways from the paper are: 
(1) We developed a new concept for how decision support systems could be designed to provide interpretability and transparency in high-stakes decisions, such as lending. It includes a globally interpretable model, along with several different types of explanation methods that might appeal to different users, including globally-consistent rules, case-based explanations, globally-consistent summary-explanations, and a visualization tool that highlights important variables and shows the full calculation of the global model.
(2) Using interpretable models in financial lending does not necessarily compromise prediction accuracy. The data for this competition was provided by FICO, whose instructions were to create a black box and explain it; they were not expecting that a globally-interpretable model could be created for these data. However, we did find such a model.
(3) Interpretability is not synonymous with sparsity. As discussed earlier, the works of \citet{Elomaa94indefense} and \cite{Freitas:2014ic} caution against models that leave out variables that might be perceived to be important (e.g., we would not want to leave out delinquency features). Thus, in our global interpretable model, we did not aim for global sparsity. We aimed for a model that decomposes into smaller models, each of which is meaningful, along with an interactive visualization tool.
(4) Summary-explanations can be useful (when not falsely described as explaining the mechanism of the black box), even for globally interpretable models. Summary-explanations can be created in a way that is consistent with the data. Our tool can be used for many decision-support systems besides financial lending.

\textbf{Limitations and Future Work:} An essential topic important to lending decisions but not handled by this work is averting bias towards protected groups. 
Fairness is a major concern in credit ratings.
However, for protected classes other than age, credit rating agencies do not have reliable or accurate demographic data for consumers or borrowers on most credit products. 
The Consumer Financial Protections Bureau (CFPB) has developed a proxy for ethnicity that uses last name and geographical location. If these data were able to be obtained or estimated, we could add constraints to the model so that it obeys a desired definition of fairness, though there is not a single definition of algorithmic fairness that is yet widely accepted, and multiple definitions of fairness conflict with each other, making them impossible to satisfy simultaneously.

Missing data is another important topic that we did not discuss in depth. Missing data is commonly imputed, but imputation of missing data can be problematic, because decisions are then made on data that can be incorrect. Another way to handle missing data is to create many models, each of which relies on a separate subset of variables. Then, if a variable is missing, we can use a model that does not rely on that variable. The approach presented in this work could be used this way, by creating a new subscale for each subset of potentially-missing variables that might appear in that subscale; once the user reports which variables are missing, we envision that the appropriate model would be automatically selected by a visualization interface. In this paper, we do not impute missing data, and individuals do not accrue points on any subscale for the \textit{values} of missing variables,  but since we have an indicator for whether certain variables are missing, our model sometimes leverages \textit{the fact that} the variable was missing to help make better predictions. Missingness is not at random, and thus whether a variable is missing can reasonably contribute to accuracy in prediction. 

A limitation that should be considered when applying our two-layer additive modeling approach to other domains is that one needs to manually decide which features constitute each subscale. This requires some knowledge of the domain. 

There are several possible direct extensions to our work. Methods like RiskSLIM \citep{UstunRu2017KDD} could make the subscale scores more interpretable by restricting to integer coefficients. Our visualization interface could be extended to be much more elaborate, like that of \citet{Ming2018} for rule-list exploration. 

Since the FICO dataset does not seem to require a black-box model for good performance, perhaps many other applications in finance also do not require a black-box model (see, e.g., \cite[][]{RudinRa2019}). The answer to this hypothesis remains unclear. However, challenges like the one initiated by FICO can help us to determine the answer to this important question.



\bibliographystyle{plainnat}
\singlespacing
\bibliography{reference}

\doublespacing
\appendix

\section{The Visualization Tool}\label{apx:vistool}

\begin{figure*}[ht]
 \includegraphics[width=1.00\linewidth]{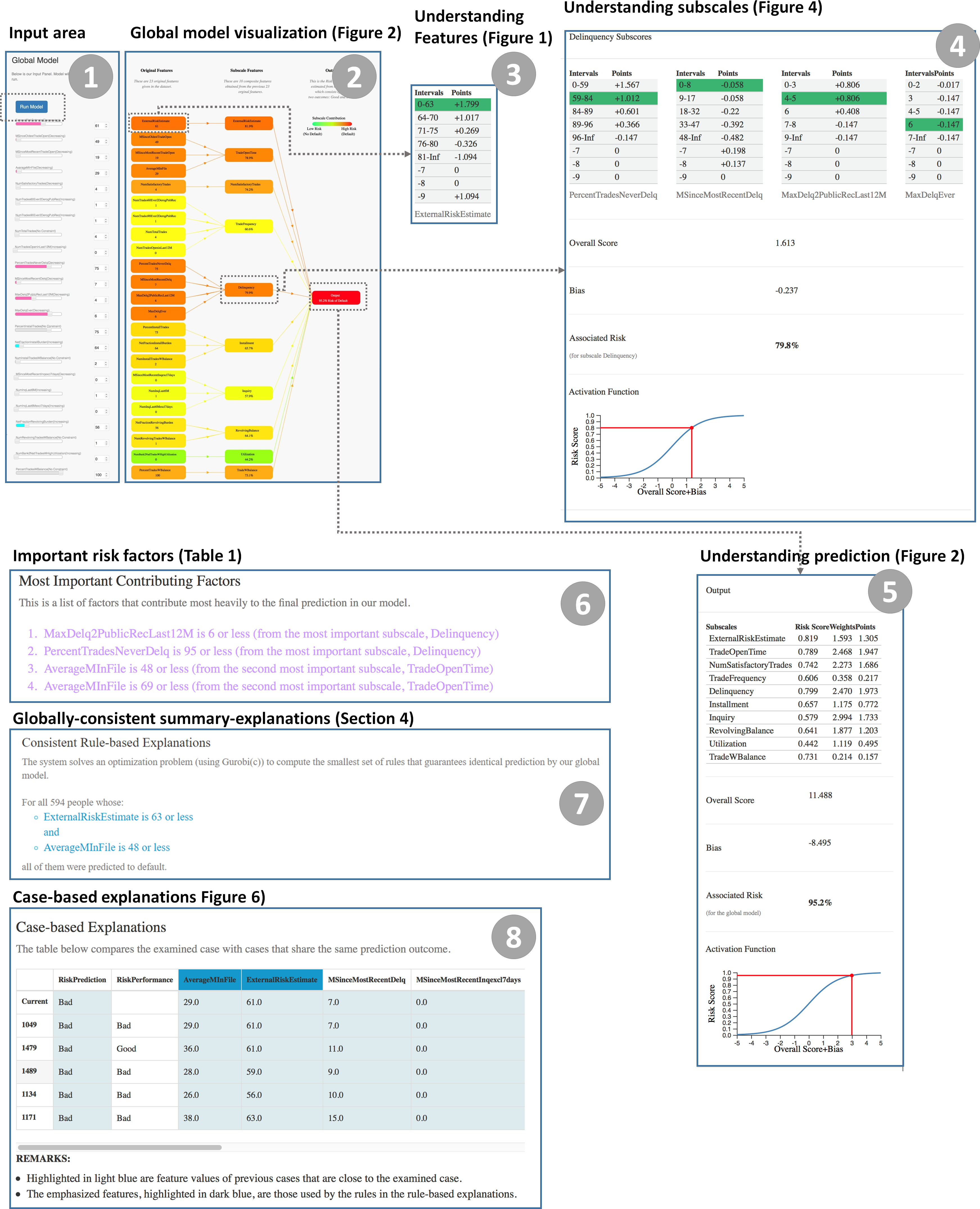}
  \caption{An overview of the visualization tool.}
  \label{fig:viztool}
\end{figure*}

Figure~\ref{fig:viztool} shows the different elements (screens) comprising the visualization tool, which can be accessed online by following this link: \url{http://dukedatasciencefico.cs.duke.edu}.

\paragraph{Input area}
Screen~1 is the input area  through which the user enters data about customer's features as well as modifies them to understand how different changes could affect prediction. For features that abide to monotonicity constraints, we used \textcolor{blue}{blue} and \textcolor{red}{red} colors to respectively denote increasing and decreasing relation between the feature's values and risk prediction. 
    
\paragraph{The global model}
Screen~2 shows the global model, represented by 3 columns.  The first (leftmost) column shows features. Clicking on one of this features opens a screen (Screen~3) which provides information about its binarization to risk factors (Section~\ref{sec:preprocessing}). The second (middle) column corresponds to subscales (Section~\ref{sec:model_design}), which aggregate one or more features, and serve as miniature risk prediction models (that is, each subscale value can be interpreted as the risk of defaulting according to its features). Clicking on each subscale shows its corresponding features and how its value is being computed for the specific prediction (see Screen~4). Finally, the right column holds the predicted output variable, that is, the risk of defaulting. Clicking on it shows the contribution of each subscale to the final prediction (see Screen~5). We used arrows to show the partition of features to subscales; we also used colors to indicate the contribution of specific features to the final prediction using a colormap that varies between green and red to indicate that a particular feature ``pulls'' the prediction toward low and high risk of defaulting. 
Screen~3 shows how the original feature values are transformed into binary values. 
Screen~4 shows the calculation of a subscale based on its features. 
Screen~5 shows the calculation of the final prediction based on the various subscales. 

\paragraph{Explanations}
    
Screen~6 shows the important risk factors (Section~\ref{sec:risk_factors} and Table~\ref{tab:ImportantFeatures}). 
Screen~7 shows a globally-consistent summary-explanation (Section~\ref{sec:summary_explanations}).
Screen~8 presents case-based explanations (Section~\ref{sub:consistent-cases} and Figure~\ref{fig:CaseBased}). 

\section{Implementation Details: Generating globally consistent rules}\label{apx:implementation_details}

While in an actual deployed system, generating rules could be done offline (since users' credit information changes over time infrequently), in our code, we wanted to allow the user the ability to interact with the system to see how explanations are generated for any possible new observation. Therefore, we wanted to limit the running time for generating explanations to provide a reasonably short response. 
To this end, we first created a database of summary-explanations using the FICO dataset. We created summary-explanations for optimal sparsity, and maximal support subject to sparsity constraints of +0, +1, and +2 of optimal. We did the same for 10,000 additional random observations. When a user clicks to generate an explanation, the database is scanned, and the sparsest relevant rule whose support is greater than 10 is returned (this value was inspired by other papers on local explanations). Otherwise, if no rules were found, then we solve the optimal sparsity optimization problem. If the support of the returned rule is greater than 10, that solution is returned. Otherwise, the maximal support optimization is solved, consecutively relaxing the constraint on the optimal sparsity as described above. If rules with sufficiently large support were not found, the procedure is repeated with minimal support set to 5 (instead of 10). Beyond this, an error message is displayed if no summary-explanations were found, because the observation is an outlier -- there is no rule characterizing it. Note that this is valuable feedback -- it tells the user of the system that the predictions of the global model cannot be compactly described near this observation, which may warrant the user to be more careful about the prediction generated by the global model near this point.

Intuitively, \texttt{OptConsistentRule} aims to provide a practical solution for generating globally-consistent rules by limiting the running time of the optimization, caching  solutions, reusing MIP solutions as basic feasible solutions for next iterations, and limiting the search space for summary-explanations whose sparsity is at most 2 from the optimal sparsity. The latter is driven by the fact that, on average, around 3 features are required to generate an optimally sparse consistent rule, and therefore adding two additional features to improve sparsity may result in rules that are too complex and not of interest to the user. 
Of course, one may make alternative design choices and tweak the above parameters.

\section{Evaluation of the Two-Layer Additive Model on the German Credit Data}\label{apx:german}

In this appendix, we briefly describe the publicly available ``German Credit Dataset''\footnote{\url{https://archive.ics.uci.edu/ml/datasets/statlog+(german+credit+data)}} and present our experimental results on this dataset.

The German Credit Dataset consists of 1,000 observations (corresponding loan applicants) and 20 features, including some categorical and some numerical features. Each loan applicant is labeled as either ``Good'' (low credit risk) or ``Bad'' (high credit risk). Unlike the FICO dataset, the German Credit Dataset is unbalanced: only $30\%$ of the observations in this dataset are labeled ``Bad'' (high credit risk). Citing the data source: ``it is worse to classify a customer as good when they are bad than it is to classify a customer as bad when they are good.'' Thus we used the weight of $5$ for the ``Bad'' class, and the weight of $1$ for the ``Good'' class in the training cost for all models.\footnote{This is achieved by duplicating the ``Bad'' observations in the training set five times.} This means that classifying a customer as good when they are bad would be penalized more severely ($5$ times more) than classifying a customer as bad when they are good. This class weighting scheme encourages a classifier to predict the minority ``Bad'' label, so that the trained classifier will not be a trivial model (which predicts ``Good'' for everyone) and will handle class imbalance well.

We trained a two-layer additive risk model on the German Credit Dataset. We used $2$ subscales: one for credit and loan information, and the other for personal information. Since this dataset is relatively small, using more than $2$ subscales tends to result in overfitting. In terms of monotonicity, 
we enforced monotonicity constraints on the two attributes ``Status of existing checking account'' and ``Savings account/bonds'' when we trained our two-layer additive risk model -- this means that if a person has more money in their checking/savings account, the credit risk predicted by our two-layer additive risk model will be lower.

Table \ref{table:numerical_results_german} shows the test AUC, average precision, and recall (at $0.5$ probability threshold) of our two-layer additive risk model on the German Credit Dataset, in comparison with other baseline models. We do not report test accuracy in this case, because test accuracy would be a misleading measure on a highly unbalanced dataset. Instead, we report recall (at $0.5$ probability threshold) of ``Bad'' observations, since it is more important to identify the bad observations than to identify the good ones (as discussed above). On this dataset, our two-layer additive risk model achieves the best AUC. In terms of average precision and recall (at $0.5$ probability threshold), our two-layer additive risk model performs comparably to the best performing model, and the performance gap between our two-layer additive risk model and the best performing model is not statistically significant (by the paired sample $t$-test).

\begin{table}[th]
\caption{Comparison of our two-layer additive risk model with baseline models, in terms of test accuracy, AUC, and average precision on the German Credit Dataset. We performed 10-fold stratified cross validation. The numbers were obtained by averaging the results over the 10 held-out test folds. Values in italics are not statistically significantly different ($p$-value greater than $0.05$ under the paired sample $t$-test) from the best performance (in bold).}
\label{table:numerical_results_german}
\centering
\begin{small}
\begin{tabular}{|l|c|l|l|l|}
\hline
Classifier                                                                           & Monotonicity?             & AUC                        & Average precision                   & Recall@.5                           \\ \hline
Two-layer additive risk model                                                        & \checkmark & $\mathbf{0.807} \pm 0.034$ & $\mathit{0.631} \pm 0.084$          & $\mathit{0.866} \pm 0.042$          \\ \hline
Single-layer additive risk model                                                     & \checkmark & $\mathit{0.802} \pm 0.036$ & $\mathit{0.622} \pm 0.083$          & $\mathit{0.860} \pm 0.044$          \\ \hline
Logistic regression                                                                  &      \xmark                     & $\mathit{0.801} \pm 0.036$ & $\mathit{0.621} \pm 0.083$          & $\mathit{0.856} \pm 0.047$          \\\hline
CART                                                                                 &   \xmark                        & $0.746 \pm 0.027$          & $0.557 \pm 0.041$                   & $\mathit{0.886} \pm 0.054$          \\ \hline
Random forest                                                                        &     \xmark                      & $0.756 \pm 0.043$          & $\mathit{0.590} \pm 0.070$          & $0.496 \pm 0.065$                   \\ \hline
Boosted trees                                                                     &  \xmark                         & $0.736 \pm 0.036$          & $0.554 \pm 0.059$                   & $0.700 \pm 0.086$                   \\ \hline
XGBoost                                                                              &     \xmark                      & $\mathit{0.798} \pm 0.025$ & $\mathit{0.631} \pm 0.055$          & $\mathit{0.840} \pm 0.044$          \\ \hline
SVM: linear                                                                          &  \xmark                         & $0.787 \pm 0.032$          & $\mathit{0.590} \pm 0.082$          & $\mathit{0.840} \pm 0.041$          \\ \hline
SVM: RBF                                                                             &  \xmark                         & $0.788 \pm 0.036$          & $\mathit{0.590} \pm 0.065$          & $\mathit{\mathbf{0.890}} \pm 0.044$ \\ \hline
\begin{tabular}[c]{@{}l@{}}NN: 2 layer, sigmoid\\ (hidden dimension 2)\end{tabular}  &     \xmark                      & $\mathit{0.804} \pm 0.035$ & $\mathit{0.623} \pm 0.086$          & $\mathit{0.866} \pm 0.042$          \\ \hline
\begin{tabular}[c]{@{}l@{}}NN: 2 layer, ReLU\\ (hidden dimension 2)\end{tabular}     &    \xmark                       & $\mathit{0.800} \pm 0.038$ & $\mathit{\mathbf{0.648}} \pm 0.077$ & $\mathit{0.843} \pm 0.044$          \\ \hline
\begin{tabular}[c]{@{}l@{}}NN: 2 layer, sigmoid\\ (hidden dimension 10)\end{tabular} &    \xmark                       & $\mathit{0.802} \pm 0.038$ & $\mathit{0.629} \pm 0.082$          & $\mathit{0.856} \pm 0.044$          \\ \hline
\begin{tabular}[c]{@{}l@{}}NN: 2 layer, ReLU\\ (hidden dimension 10)\end{tabular}    &    \xmark                       & $0.773 \pm 0.033$          & $\mathit{0.597} \pm 0.060$          & $0.736 \pm 0.062$                   \\ \hline
NN: 4 layer, sigmoid                                                                 &      \xmark                     & $\mathit{0.800} \pm 0.036$ & $\mathit{0.624} \pm 0.083$          & $\mathit{0.823} \pm 0.081$          \\ \hline
NN: 4 layer, ReLU                                                                    &  \xmark                         & $0.737 \pm 0.062$          & $0.555 \pm 0.066$                   & $0.570 \pm 0.070$                   \\ \hline
NN: 8 layer, sigmoid                                                                 &  \xmark                         & $\mathit{0.801} \pm 0.048$ & $\mathit{0.629} \pm 0.100$          & $0.696 \pm 0.073$                   \\ \hline
NN: 8 layer, ReLU                                                                    &     \xmark                      & $0.739 \pm 0.060$          & $0.542 \pm 0.109$                   & $0.613 \pm 0.093$     \\ \hline             
\end{tabular}
\end{small}
\end{table}


\bibliographystyle{ACM-Reference-Format}
\bibliography{reference}

\end{document}